\documentclass[preprint,12pt]{elsarticle}

\usepackage{graphicx}
\usepackage{amssymb}

\usepackage{lineno}

\usepackage{tipa}

\usepackage{textcomp}
\usepackage{multirow}
\usepackage{color}
\usepackage{makecell}

\usepackage{float}
\usepackage{graphicx}
\usepackage{hhline}
\usepackage{tabularx}
\usepackage{amsmath,amsfonts,amssymb,amsthm}
\usepackage{mathtools}
\DeclarePairedDelimiter{\norm}{\lVert}{\rVert}
\usepackage{textcomp}
\usepackage[table]{xcolor}
\usepackage{caption}
\usepackage{subcaption}

\journal{Journal Name}

\begin{document}

\begin{frontmatter}





\title{Large-Scale Acoustic Characterization \\of Singaporean Children’s English Pronunciation}


\author[]{Yuling Gu\corref{cor1}}
\ead{yulinggu@uw.edu}

\cortext[cor1]{Corresponding authors}

\author[]{Nancy F. Chen\corref{cor1}}
\ead{nancychen@alum.mit.edu}

\begin{abstract}
In this work\let\thefootnote\relax\footnotetext{Yuling Gu is at University of Washington when this work is submitted to arXiv, but the majority of the work was completed while at New York University.
Nancy F. Chen is currently with the Institute for Infocomm Research,
A*STAR.}, we investigate pronunciation differences in English spoken by Singaporean children in relation to their American and British counterparts by conducting Kmeans clustering and Archetypal analysis on selected vowel pairs and approximants. Given that Singapore adopts British English as the institutional standard due to historical reasons, one might expect Singaporean children to follow British pronunciation patterns. Indeed, Singaporean and British children are more similar in their production of syllable-final rhotic approximant /\textipa{\*r}/ -- they do not lower their third formant (F3) nearly as much as American children do, suggesting a lack of rhoticity. Interestingly, we also observe that Singaporean children present similar patterns to American children when it comes to their fronting of vowels as demonstrated across TRAP$-$BATH split vowels, /\textipa{\ae}/, /\textipa{E}/, /\textipa{A}/, /\textipa{O}/, /\textipa{i}/ and /\textipa{I}/. Singaporean children's English also demonstrated characteristics that do not resemble any of the other two populations. We observe that Singaporean children's vowel height characteristics are distinct from both that of American and British children. In tense and lax vowel pairs, /\textipa{u}/ and /\textipa{U}/, as well as /\textipa{i}/ and /\textipa{I}/, we also consistently observe that the difference is less conspicuous for Singaporean children compared to the other speaker groups.  Further, while American and British children demonstrate lowering of F1 and F2 formants in transitions into syllable-final /\textipa{l}/s, a wide gap between F2 and F3 formants, and small difference between F1 and F2 formants, all of these are not exhibited in Singaporean children's pronunciation. These findings point towards potential sociolinguistic implications of how Singapore English might be evolving to embody more than British pronunciation characteristics. Furthermore, these findings also suggest that Singapore English could be have been influenced by languages beyond American and British English, potentially due to Singapore's multilingual environment.

\end{abstract}


\end{frontmatter}



\section{Introduction}
\label{sec1}

English varieties in the world can be represented in the form of three concentric circles -- inner circle (e.g. US, UK), outer circle (e.g. Singapore, India), and expanding circle (e.g. China, Russia) \cite{kachru1982EnglushAcrossCult}. The inner circle contains Anglo Englishes whereas the outer circle contains `New Englishes' of which the spread of English to those regions occurred through the process of historical colonization. 
Extensive work has been done to investigate American English, including acoustic, phonetic or sociolinguistic studies \cite{Kuo2013FormantTI, Clopper2007FreeCO, chen2009channel, labov2006, evanini2008clusterNorthAmE} and work using machine learning to automatically find pronunciation patterns \cite{chen2014mlPhoneticTransf, nancythesis, Tauberer2009AmEVowel}. 
There is also much work on studying different varieties of British English in terms of phonetics and prosody, including \cite{henton1983ChangesRPVowel, Grabe2002IV, wells1999RPChangingScene, Low1998ACS}.
Further, these two inner circle English pronunciations have often been compared to each other \cite{khan2019CompareBE&AE, gomez2009CompareBE&AE}.

By contrast, investigations on English spoken by groups in the outer circle (e.g., Indian English, Singapore English) has received much less attention. For Singapore English, there has been literature providing analysis at length at the syntactic level (e.g. \cite{Alsagoff1998grammar}) and at the semantic level (e.g. \cite{wong2004ParticlesSG}). Analysis from a phonological perspective mainly focused on patterns from stress, rhythm and intonation (e.g. \cite{Ling2000quantitativeSG, grabe2003Intonation, yeow_1986_rhythm}), yet few have examined speech acoustic characteristics. Previous phonological analysis in this direction have either been based on anecdotal evidence (e.g. \cite{deterding1994teaching, foley_new_1988}) or have been limited in scale due to the lack of available large-scale corpora and the limited number of speakers recruited for the experiments. For example, the National Institute of Education Corpus of Spoken Singapore English \cite{niecsse} consists of five-minute long interviews from 31 female and 15 male speakers; \cite{deterding2007SgE}'s phonological analysis was mainly based on a one-hour recording of a single female speaker.
Previous work outlined some distinctive phonological features of Singapore English \cite{tan2012EnglishInSG}; for instance, /\textipa{\ae}/ in British Received Pronunciation (RP) and general American pronunciation are more likely to be acoustically realized as other vowels like /\textipa{E}/ in Singapore English. \cite{deterding2007SgE} gave a comprehensive description of the features of Singapore English by analyzing various phonemes in speech collected from one female undergraduate student. However, till date, there has been no large-scale experiment to quantify these  observations. Furthermore, all such work focuses on adult speech, while studies on child speech, which have important applications such as computer-assisted language learning, is limited, if any. 

In this work, we present a large-scale analysis to acoustically quantify the characteristics of Singaporean children's English pronunciations for various vowel pairs and approximants. The speaker number and utterance number in this study are at least an order of magnitude greater than past work such as \cite{niecsse, deterding2007SgE}.
This work extends and expands our prior work \cite{Gu2020CharacterizationOS} to include more experiments, analyses and discussions left out in the conference version. In particular, we expand our analyses to cover more vowels and also include studies of different approximants. This work can be seen as expanding and developing the directions briefly outlined in our earlier abstracts \cite{Gu2019AcousticWiNLP, Gu2019LargescaleAC} for the analysis of vowels and \cite{Gu2019AcousticCO} for the analysis of approximants respectively.
Our findings could have various applications in areas such as native language identification \cite{malmasikeelan17shared}, automatic speech recognition \cite{karenthesis, livescu00modelnonNaiveASR}, self-correction behavior in vowel production \cite{Bakst2019l1l2vowel}, and be applied to improve pronunciation modeling in computer-assisted language learning \cite{tong2014subspace, lee2016personalized, chen2016pronounciationscore,ke2020speecheval}. 

\section{Experimental Design}
\label{sec2}
\subsection{Speech Corpora}
\label{subsec1}

Read speech was collected from American children (140 speakers, 43,406 utterances), British children (95 speakers, 32,542 utterances) and Singaporean children (193 speakers, 34,457 utterances). The age range is 6-13 years old and the gender ratio is balanced. The reading material were customized for each of the three populations, and consists of sentences from TIMIT \cite{timit}, PF-STAR \cite{pf-star, batliner2005pf-star}, GMU Speech Accent Archive \cite{gmu} and carefully designed sentences containing minimal pairs and words that elicit possible acoustic and pronunciation differences across speakers and speaker populations. All three corpora were designed to be phonetically balanced, and in part designed according to the considerations laid out in \cite{chen2016singakids,chen2015icall, icall}.

\subsection{Acoustic Features}
\label{subsec2}
The Praat \cite{boersma2001praat} software was used to extract acoustic features from the utterances on a per phoneme basis. We force-aligned the utterances to produce time boundaries, a small sample of which were manually inspected to ensure that they are accurate within 50ms of the actual boundaries. The acoustic features used included estimates of the first four formants, F0 and phoneme duration. These features were sampled at a 10ms time step and averaged for each phoneme to give us a representation of its acoustics characteristics. Using these features, unsupervised clustering approaches were used to explore patterns in the data, detailed acoustic analysis were used as a follow-up to gain phonological insights, and conjectures of articulatory gestures were made according to acoustic phonetics knowledge \cite{Stevens:98}. 

\subsection{Archetypal Analysis}
\label{subsec3}
Most algorithms in unsupervised clustering such as k-means \cite{macqueen1967Kmeans} use centroids to conduct cluster analysis. 
In this work, motivated by multilingual and multicultural influence of Singapore English, we adopt Archetypal analysis \cite{cutler1994archetypal} to investigate how American and British (inner circle English) pronunciations might serve as anchoring Archetypal references to characterize Singapore English (outer circle English). All experiments were also conducted on k-means. As both approaches show similar trends, we will focus on the results for achetypal analysis.

Archetypal analysis represent each data point in a data set as a combination of characteristic ``archetypes'' (pure types) \cite{cutler1994archetypal}. Given a set of multivariate data,  $\{ \mathbf{x_i}, i = 1, ..., n \}$, where each $\mathbf{x_i}$ is a vector of length $m$, we seek vectors $\mathbf{z_1}, ..., \mathbf{z_p}$ of length $m$ that form the Archetypal extremes. The vectors $\mathbf{z_1}, ..., \mathbf{z_p}$ are defined as 
\begin{equation}
\mathbf{z_k} = \sum_{j} \beta_{kj}\mathbf{x_j} , k = 1,...,p
\end{equation}
where $\beta_{ki} \geq 0, \sum_{i}\beta_{ki} = 1 $, and we define $\{\alpha_{ik}\}, k = 1,...,p$ to minimize the following expression
\begin{equation}
\norm[\Bigg]{\mathbf{x_i} - \sum_{k = 1}^{p} \alpha_{ik}\mathbf{z_k}}^{2}
\end{equation}
where $\alpha_{ik} \geq 0, \sum_{k}\alpha_{ik} = 1 $. The archetypes are defined as vectors $\mathbf{z_1}, ..., \mathbf{z_p}$ that minimize D, where D is the sum of squares of distances from each data vector $ \mathbf{x_{i}} $ to the convex hull formed by the $\mathbf{z_1}, ..., \mathbf{z_p}$ vectors.


\begin{equation}
D = \sum_{i}\norm[\Bigg]{\mathbf{x_i} - \sum_{k = 1}^{p} \alpha_{ik}\mathbf{z_k}}^{2}
\end{equation}
Archetypal analysis applies an alternating minimizing algorithm to a nonlinear least squares problem. 
We experiment with this approach using an R package for Archetypal analysis that has been documented in  \cite{eugsterAndCo2009archetypes} and  \cite{eugster2010archetypes}. For the following analysis we present, we experimented with various values for $p$ and chose $p$ = 2 using ``elbow criterion'' on the residual sum of squares \cite{eugsterAndCo2009archetypes}.\\

\subsection{Data Processing}
\label{subsec4}
For the purposes of our analysis, with all the phoneme tokens extracted from our corpora, we first excluded silent tokens (labeled `SIL'), non-silent tokens with F1, F2 or F3 undefined, as well as non-silent tokens with duration greater than 1s. The remaining tokens are then averaged on a per speaker and per phoneme basis, giving one data point for each speaker, each phoneme. To analyze syllable-final and syllable-initial /\textipa{l}/ and /\textipa{\*r}/, we used the CMU Pronouncing Dictionary version 0.6 augmented with syllable boundaries (syllabified CMU) \citep{Bartlett2009OnTS} for syllabification. 

Using the Nordstroem and Lindblom model \cite{nordstrom_1975_norm}, we computed scaling factors for normalization within (e.g. age, gender) and across speaker groups (population). The model was adopted for normalization across the languages based on speaker groups to account for anatomical differences in vocal tract length. This model estimate the total length of subject's vocal tract using average of F3 in vowels with F1 greater than 600Hz. In our adaptation of the model, we determined vowels by referencing \citep{Jurafsky2021}.

\section{Overall formant space}

\begin{figure}
	\centering
	\includegraphics[width=\linewidth]{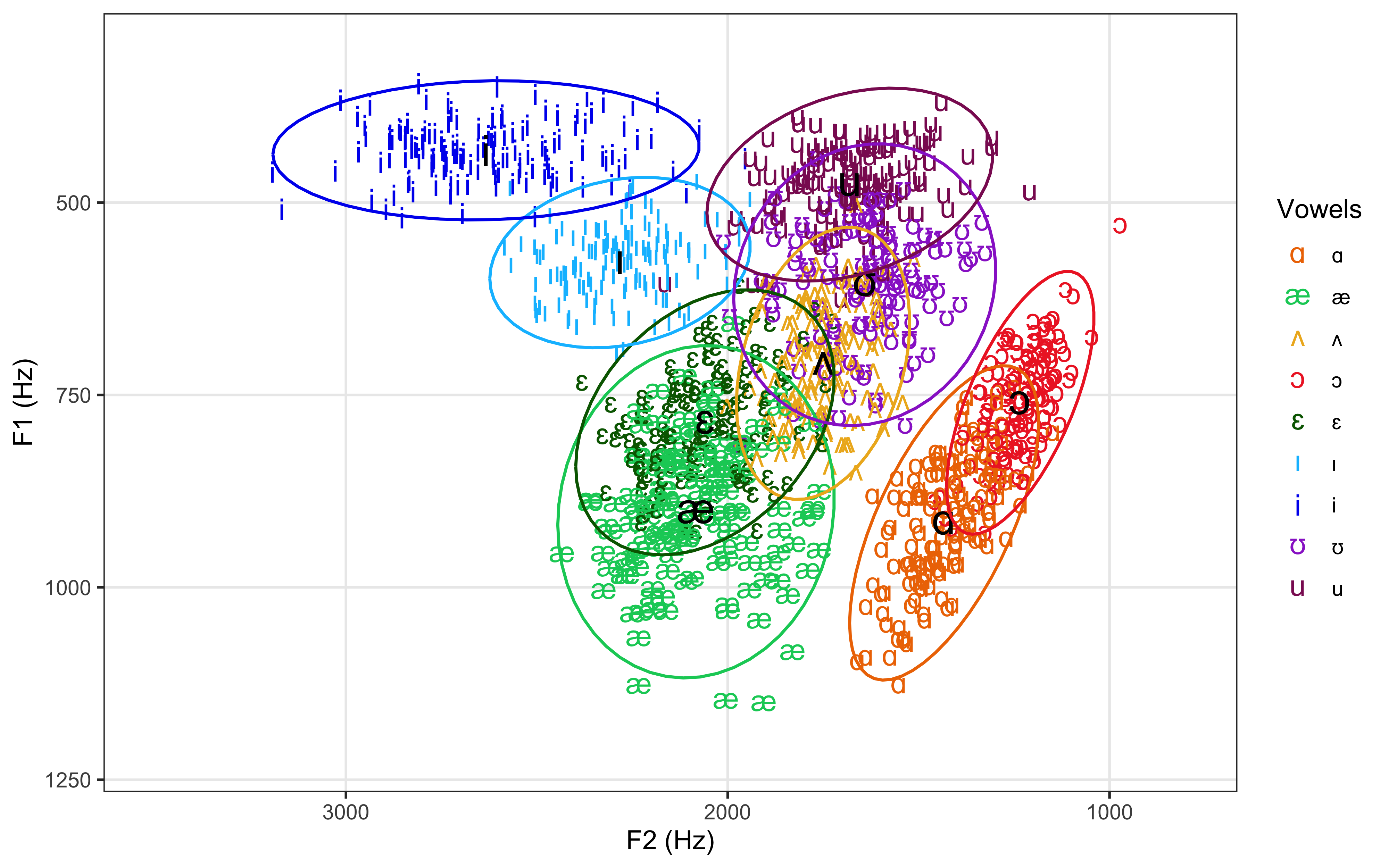}
    \caption{Overall formant space for American children}
    \label{fig:AmE_overall_formantspace}
\end{figure}

\begin{figure}
	\centering
	\includegraphics[width=\linewidth]{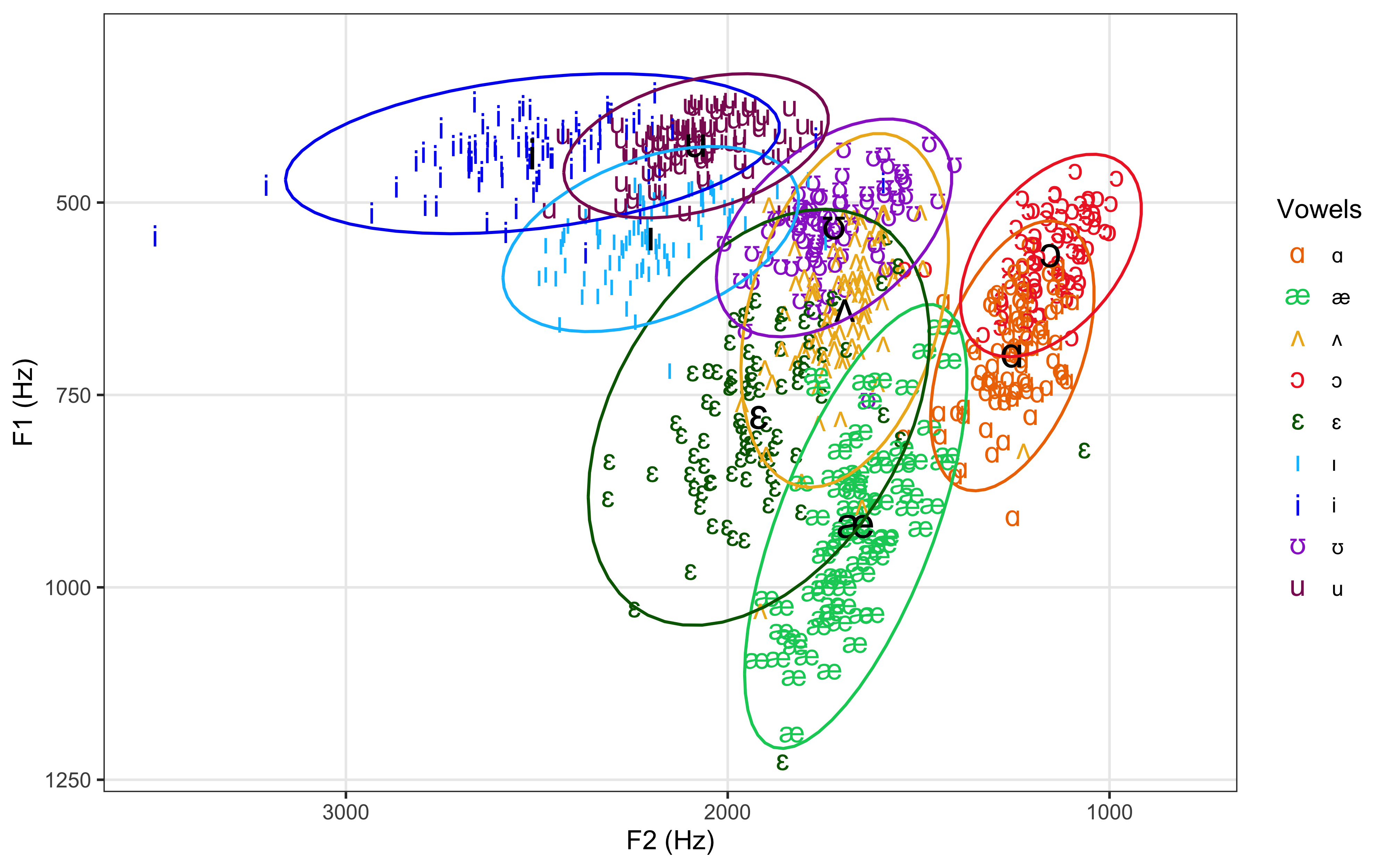}
    \caption{Overall formant space for British children}
    \label{fig:BE_overall_formantspace}
\end{figure}

\begin{figure}
	\centering
	\includegraphics[width=\linewidth]{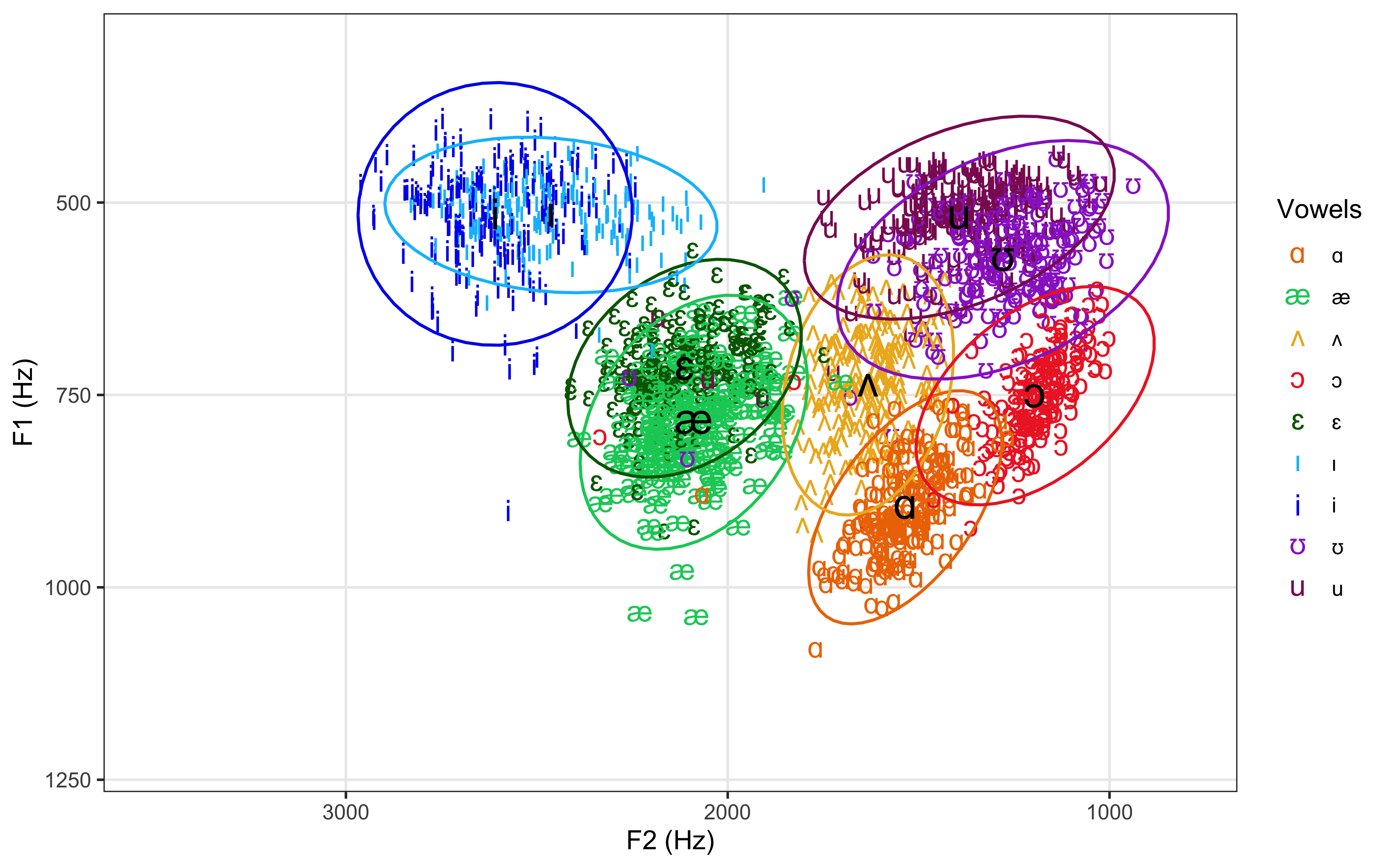}
    \caption{Overall formant space for Singaporean children}
    \label{fig:SG_overall_formantspace}
\end{figure}

We present an overview of the formant space for American, British and Singaporean children in Figures \ref{fig:AmE_overall_formantspace}, \ref{fig:BE_overall_formantspace} and \ref{fig:SG_overall_formantspace} respectively. These vowel plots suggest some trends in how the three speaker groups  may differ in their realization of the different vowels, including:
\begin{itemize}
    \item American and Singaporean children both have /\textipa{\ae}/ and  /\textipa{E}/ vowels that are greatly overlapping in the formant space. British children's /\textipa{\ae}/ and  /\textipa{E}/ vowels also show some overlap in F1 and F2 features.
    \item All three populations demonstrate some overlapping of /\textipa{A}/ and /\textipa{O}/ in the formant space (albeit at varying degrees). 
    \item American and Singaporean children both have /\textipa{u}/ and /\textipa{U}/ that are overlapping in the formant space, though this is not the case for British children.
    \item Singaporean children show great overlap in their /\textipa{i}/ and /\textipa{I}/ vowels. Such great overlap is not observed in the other two speaker groups for /\textipa{i}/ and /\textipa{I}/.
\end{itemize}

We examine and analyze vowels produced by the three speakers in greater detail in Sections \ref{sec:trap_bath} to \ref{sec:sec_long_short_i}. 

\section{Trap-bath split}
\label{sec:trap_bath}
TRAP$-$BATH split is a vowel split that is well-known in UK (including RP) \cite{wells1982AccentsOfEnglish}, where vowels in words such as \textit{glass, laugh, dance, can't} are pronounced as the [\textipa{A}] phone instead of [\textipa{ae}] as in \textit{trap, cab, mad}. Such splitting is not typically observed in American English \cite{wells1982AccentsOfEnglish}. In this section, we examine how Singaporean, American, and British children might produce TRAP$-$BATH split vowels, where [\textipa{A}] and [\textipa{ae}] phones are the different realizations of such vowels. For the rest of this paper, we refer to vowels that could turn into the [\textipa{A}] phone when TRAP$-$BATH split is present as [\textipa{A}] vowels and those that are realized as the [\textipa{ae}] phone as [\textipa{ae}] vowels.

\subsection{Unsupervised clustering}
\label{subsec:trap_bath_unsup}
Table \ref{tab:TRAP_BATH_archetypal_F1F2} shows the clustering results of Archetypal analysis using F1 and F2. We perform one clustering experiment per speaker group on TRAP$-$BATH split vowels from that group. For each phone (i.e. [\textipa{ae}] and [\textipa{A}]), we show the percentage of tokens from that category that get clustered into each of the cluster groups. For each row, the two numbers add up to 1.0 (stands for 100\%). This illustrates the proportion of tokens that gets grouped into Group 1 and Group 2 respectively. Majority of Singaporean children's [\textipa{ae}] and [\textipa{A}] vowels in the TRAP$-$BATH split are largely ($>60\%$ of each of these vowels) grouped to one cluster, suggesting that Singaporean children may produce these vowels with less acoustic distinction in terms of the formant estimates. In contrast, more distinctive clusters were observed for American and British children's TRAP$-$BATH split vowels. Results from Kmeans clustering in Table \ref{tab:TRAP_BATH_kmeans_F1F2} similarly show that American and British children's TRAP$-$BATH split vowels give cleaner clusters. In the next subsection, we investigate if cleaner clusters indeed reflect the split through formant space analysis. 

\begin{table}[t]
  \begin{center}
    \small\addtolength{\tabcolsep}{-1pt}
	\begin{tabular}{|c|c|c|}
    \hline
    \multicolumn{3}{|c|}{Singaporean Children} \\
    \hline
   	\multirow{2}{*}{Phone} & Group1 & Group2 \\
     &  (625, 1769) & (937, 2323) \\
   	\hline
   	[\textipa{\ae}] & 0.391 & \textbf{0.609} \\
   	{[\textipa{A}]}  & 0.385 & \textbf{0.615}  \\
   	\hline
   	\multicolumn{3}{c}{} \\
   	\hline
   	\multicolumn{3}{|c|}{American Children} \\
   	\hline
   	\multirow{2}{*}{Phone} & Group1 & Group2 \\
   	 &  (1081, 1780) & (694, 2426) \\
   	\hline
   	[\textipa{\ae}] &   \textbf{0.836}  & 0.164 \\
   	{[\textipa{A}]} & 0.221 &  \textbf{0.779} \\
   	\hline
   	\multicolumn{3}{c}{} \\
   	\hline
   	\multicolumn{3}{|c|}{British Children} \\
   	\hline
   	 \multirow{2}{*}{Phone} & Group1 & Group2 \\
   	  &  (1112, 1896) & (698, 1346) \\
   	\hline
   	[\textipa{\ae}] &   \textbf{0.707}  & 0.293 \\
   	{[\textipa{A}]} & 0.349 &  \textbf{0.651} \\
   	\hline
    \end{tabular}
    \caption{Archetypal analysis using \textit{F1}(Hz), \textit{F2}(Hz) estimates for Singaporean, American and British children for TRAP$-$BATH split [\textipa{\ae}] and [\textipa{A}] vowels. Archetypal extreme points for each cluster are in the format (F1, F2).}
    \label{tab:TRAP_BATH_archetypal_F1F2}
  \end{center}
    
\end{table}

\begin{table}[t]
  \begin{center}
    \small\addtolength{\tabcolsep}{-1pt}
	\begin{tabular}{|c|c|c|}
    \hline
    \multicolumn{3}{|c|}{Singaporean Children} \\
    \hline
   	\multirow{2}{*}{Phone} & Group1 & Group2 \\
     &  (785, 1961) & (806, 2178) \\
   	\hline
   	[\textipa{\ae}] & 0.339 &  \textbf{0.661} \\
   	{[\textipa{A}]}  & \textbf{0.630}  & 0.370 \\
   	\hline
   	\multicolumn{3}{c}{} \\
   	\hline
   	\multicolumn{3}{|c|}{American Children} \\
   	\hline
   	\multirow{2}{*}{Phone} & Group1 & Group2 \\
   	 &  (918, 1974) & (850, 2271) \\
   	\hline
   	[\textipa{\ae}] &   \textbf{0.800}  & 0.200 \\
   	{[\textipa{A}]} & 0.350 &  \textbf{0.650} \\
   	\hline
   	\multicolumn{3}{c}{} \\
   	\hline
   	\multicolumn{3}{|c|}{British Children} \\
   	\hline
   	 \multirow{2}{*}{Phone} & Group1 & Group2 \\
   	  &  (983, 1755) & (840, 1503) \\
   	\hline
   	[\textipa{\ae}] &   \textbf{0.756}  & 0.244  \\
   	{[\textipa{A}]} & 0.277 &  \textbf{0.723} \\
   	\hline
    \end{tabular}
    \caption{Kmeans clustering using \textit{F1}(Hz), \textit{F2}(Hz) estimates for Singaporean, American and British children for TRAP$-$BATH split [\textipa{\ae}] and [\textipa{A}] vowels. Cluster centriods for each cluster are in the format (F1, F2).}
    \label{tab:TRAP_BATH_kmeans_F1F2}
  \end{center}
    
\end{table}

\subsection{Acoustic Analysis and Characterization}
\label{subsec:trapbath_acoustic}
 \subsubsection{F1 F2 formant space}
\begin{figure}[t]
	\includegraphics[width=1.16\linewidth]{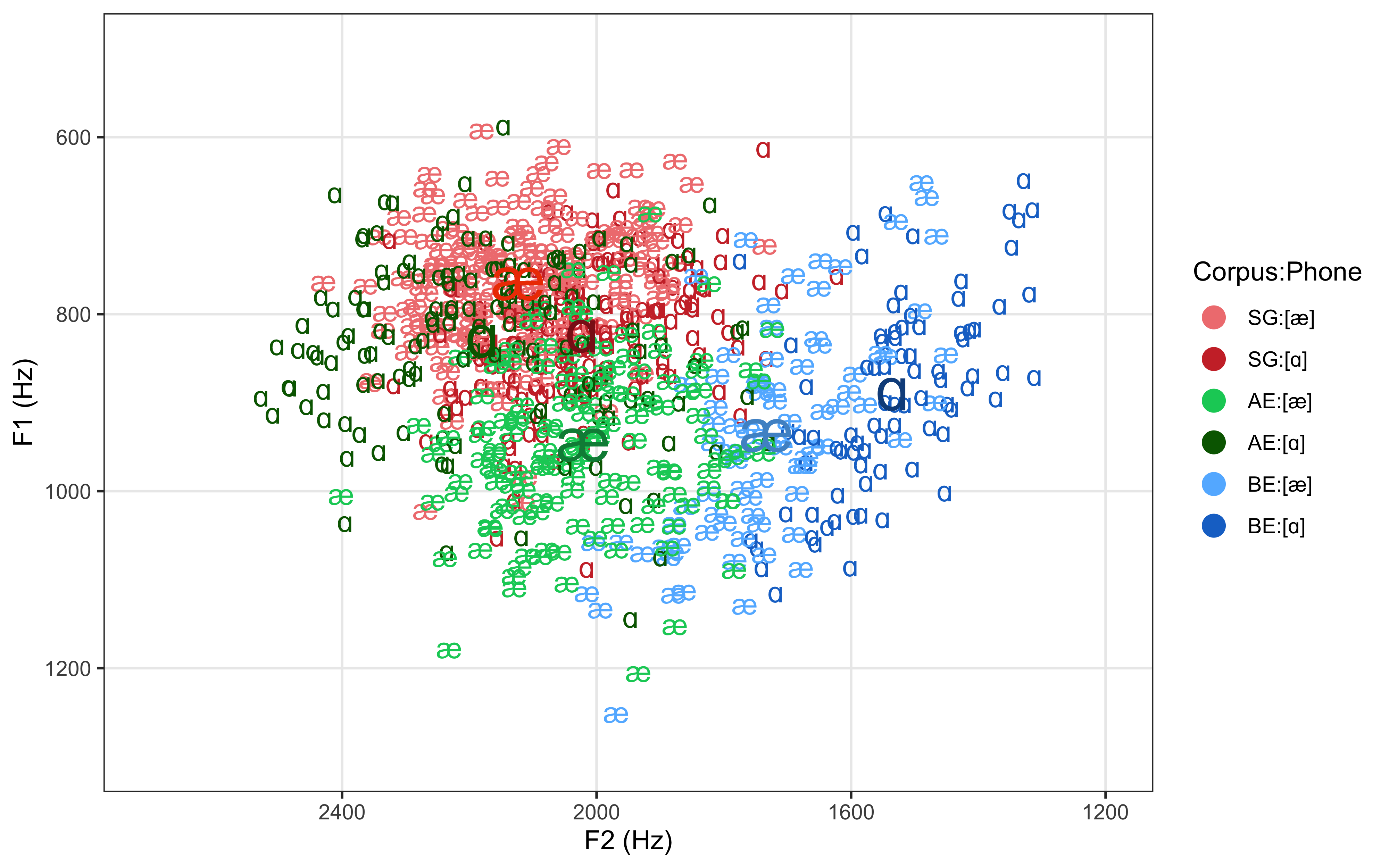}
	\caption{Singaporean and American children are more similar in terms of \textit{F1} and \textit{F2} estimates for TRAP$-$BATH split vowels compared to British children. /\textipa{ae}/ phonemes that could turn into [\textipa{A}] in TRAP$-$BATH split are labeled as [\textipa{A}] while those that are realized as [\textipa{ae}] are labeled as [\textipa{ae}]. Smaller points: individual speaker's mean \textit{F1} and \textit{F2}; larger points: group means for the speaker groups.}
    \label{fig:AE_AA_comparison}
\end{figure}

 We present the mean and standard error for F1 and F2 estimates in Table \ref{tab:AE_AA_table} and visualize this data on a per speaker level in Figure \ref{fig:AE_AA_comparison}: we observe some overlap between the American and Singaporean populations, which are more fronted than British pronunciations (higher F2). We observe that TRAP$-$BATH split vowels produced by Singaporean children generally have lower F1 values compared to American and British speakers. We then further analyzed this effect in detail in terms of F1 and F2 separately.

\begin{table}[t]
  \begin{center}
  \small\addtolength{\tabcolsep}{-1pt}
  \begin{tabular}[\linewidth]{|c|c|c|c|c|c|}
	\hline
	\bf Corpus & Phone & F1 mean & F1 se & F2 mean & F2 se \\
	\hline
      SG & [\textipa{\ae}] & 766 &  5.03 & 2123 & 9.13 \\
      & [\textipa{A}] & 825 & 5.09 & 2022 & 8.89 \\ 
      \hline
      AE & [\textipa{\ae}] & 948 & 7.56 & 2021 & 11.5 \\
      & [\textipa{A}] & 831 & 8.01 & 2179 & 15.9 \\
      \hline
      BE & [\textipa{\ae}] & 937 & 13.5 & 1733 & 14.3 \\
      & [\textipa{A}] & 891 & 12.0 & 1535 & 12.4  \\
	\hline
  \end{tabular}
  \caption{Mean and standard error (se) for each speaker group for TRAP$-$BATH split vowel formants.}
  \label{tab:AE_AA_table}
  \end{center}
\end{table}

\subsubsection{F1 formant estimates}
For TRAP$-$BATH split vowels that become [\textipa{ae}] under the split, Singaporean children have the lowest F1 (\textit{M} = 766), American children have statistically significantly higher F1 (\textit{M} = 948); British children (\textit{M} = 937) show similar trends to Americans. For vowels that could turn into [\textipa{A}] when the split is present, Singaporean children similarly have the lowest F1 (\textit{M} = 825), American children have slightly higher F1 (\textit{M} = 831), and British children have the highest F1 (\textit{M} = 891). A one-way ANOVA demonstrated that these differences are statistically significant, \textit{F}(2, 411) = 207.1, \textit{p} $<$ 0.001 for [\textipa{ae}] vowels under the split, and \textit{F}(2, 412) = 17.3, \textit{p} $<$ 0.001 for vowels that could turn into [\textipa{A}] under the split. We further investigate which pairs of these three populations are significantly different from each other. Thus we followed up our ANOVA test with a post hoc Tukey's HSD Test: For vowels that are realized as [\textipa{ae}] under the split, in terms of F1, American and Singaporean children differ significantly at \textit{p} $<$ 0.001; British and Singaporean children differ significantly at \textit{p} $<$ 0.001. However, American and British children are not significantly different from each other. For vowels that turn into [\textipa{A}] under the split, British children are significantly different from the other two groups in terms of F1 ($p  <  0.001$), but American and Singaporean children are not significantly different from each other. Therefore one significant difference for TRAP$-$BATH [\textipa{ae}] vowels produced by the three populations is that those produced by Singaporean children have significantly lower F1 formant estimates compared to the other two populations. For the [\textipa{A}] vowels, those produced by both Singaporean and American children have significantly lower F1 formant estimates compared to British children. The articulatory implication is that Singaporean children exhibit a higher tongue height when producing both TRAP$-$BATH split vowels, and American children exhibit this trait when producing the [\textipa{A}] TRAP$-$BATH split vowel. 

\subsubsection{F2 formant estimates}
For vowels that are realized as [\textipa{ae}] under the split, British children show the lowest F2 (\textit{M} = 1733), American children show higher F2 (\textit{M} = 2021), and Singaporean children show the highest F2 (\textit{M} = 2123). A one-way ANOVA shows that these differences are statistically significant, \textit{F}(2, 411) = 258, \textit{p} $<$ 0.001.
For vowels that turn into [\textipa{A}] under the split, British children again show the lowest F2 (\textit{M} = 1535), Singaporean children show higher F2 (\textit{M} = 2022), and American children show the highest F2 (\textit{M} = 2179). A one-way ANOVA shows that these differences are statistically significant, \textit{F}(2, 412) = 518.6, \textit{p} $<$ 0.001. To investigate which pairs of the three speaker groups are significantly different from each other, we followed up our ANOVA test with a post hoc Tukey's HSD Test which show that all pairwise comparisons are significantly different (\textit{p} $<$ 0.001). Such findings show that in terms of articulatory implications, Singaporean and American children's productions of TRAP$-$BATH vowels are similarly more fronted, compared to British speakers.

\subsection{Summary: Singaporean children's TRAP-BATH split trends are more similar to American children and show less clear TRAP-BATH split compared to British children}
\label{subsec:trap_bath_overall}

Compared to British children, Singaporean and American children are more similar in fronting their TRAP$-$BATH vowels to produce something closer to [\textipa{\ae}] rather than [\textipa{A}]. Singaporean and American children's production of the TRAP$-$BATH vowels also differ from British children by having acoustic features that reflect a higher tongue position. Singaporean children demonstrate a higher tongue position for both [\textipa{\ae}] and [\textipa{A}] TRAP$-$BATH vowels  compared to British children while American children demonstrate the same trait for [\textipa{A}] vowels. Thus, Singaporean and American children not showing as much TRAP$-$BATH split distinction as the two vowels are produced with similar frontness that is more characteristic of [\textipa{\ae}] than [\textipa{A}].  
These articulatory features agree with our clustering results in suggesting that Singaporean children do not show TRAP$-$BATH split. The formant analysis also further clarified that while American children's TRAP$-$BATH vowels fall into two clusters, they are all produced with acoustic features like that of [\textipa{\ae}]. For American children, TRAP$-$BATH split vowels that could be changed to the back vowel [\textipa{A}] when the split is present are instead articulated with even a fronter position (higher F2) than [\textipa{\ae}], and thus pronounced like [\textipa{\ae}],  reaffirming the traditional knowledge that most Americans do not exhibit TRAP$-$BATH split. Our formants analysis show that British children's TRAP$-$BATH split exhibit more acoustic differences compared to the other two groups, and reaffirm the clustering results that they show TRAP$-$BATH split since vowels that could change into [\textipa{A}] in the split are indeed produced with lower F2 like a back vowel ($M$ = 1535), compared to those that could stay as the front vowel [\textipa{\ae}] with higher F2 ($M$ = 1733).


\section{/\textipa{\ae}/ vs. /\textipa{E}/}
\label{sec:ae_eh}
/\textipa{E}/ is a mid-low front vowel. When compared to the /\textipa{\ae}/ phoneme, the /\textipa{E}/ phoneme has slightly higher F2 and lower F1 estimates \cite{Stevens:98}. Therefore, any fronting of /\textipa{\ae}/, resulting in a higher F2, could lead to potential confusion with /\textipa{E}/. Having observed higher F2 of Singaporean and American children in their TRAP$-$BATH split vowels compared to British children (across realizations as [\textipa{\ae}] and [\textipa{A}] phones), we further examine how production of /\textipa{\ae}/ and /\textipa{E}/ phonemes might differ across the three speaker groups.

\subsection{Unsupervised Clustering}
\label{subsec:ae_eh_unsup}
We perform per-speaker-group clustering across /\textipa{\ae}/ and /\textipa{E}/ vowels from each group  using F1 and F2 estimates. Clustering results using Archetypal analysis and Kmeans clustering are shown in Tables \ref{tab:AE_EH_archetypal_clustering_F1F2} and \ref{tab:AE_EH_kmeans_clustering_F1F2}. The results from the two clustering methods agree, indicating that /\textipa{\ae}/ and /\textipa{E}/ vowels produced by the British children are largely ($>$ 90 \% for Archetypal analysis, $>$ 80 \% for Kmeans clustering) grouped into two clean, distinctive clusters, whereas such distinction is less clear cut for these vowels produced by Singaporean and American children.

\begin{table}[t]
  \begin{center}
    \small\addtolength{\tabcolsep}{-1pt}
	\begin{tabular}{|c|c|c|}
    \hline
    \multicolumn{3}{|c|}{Singaporean Children} \\
    \hline
   	\multirow{2}{*}{Phoneme} & Group1 & Group2 \\
     &  (886, 2337) & (604, 1845) \\
   	\hline
   	/\textipa{\ae}/ & \textbf{0.651} & 0.349  \\
   	/\textipa{E}/ & 0.448  & \textbf{0.552} \\
   	\hline
   	\multicolumn{3}{c}{} \\
   	\hline
   	\multicolumn{3}{|c|}{American Children} \\
   	\hline
   	\multirow{2}{*}{Phoneme} & Group1 & Group2 \\
   	 &  (1013, 2318) & (665, 1812) \\
   	\hline
   	 /\textipa{\ae}/ & \textbf{0.693} & 0.307 \\
   	 /\textipa{E}/ & 0.371 & \textbf{0.629} \\
   	\hline
   	\multicolumn{3}{c}{} \\
   	\hline
   	\multicolumn{3}{|c|}{British Children} \\
   	\hline
   	 \multirow{2}{*}{Phoneme} & Group1 & Group2 \\
   	  &  (1019, 1562) & (697, 1999) \\
   	\hline
   	/\textipa{\ae}/ &  \textbf{0.928}  & 0.072 \\
   	/\textipa{E}/ & 0.060 &  \textbf{0.940} \\
   	\hline
    \end{tabular}
    \caption{Archetypal analysis using \textit{F1}(Hz), \textit{F2}(Hz) estimates of /\textipa{\ae}/ and /\textipa{E}/from Singaporean, American and British children.  Archetypal extreme points for each cluster are in the format (F1, F2).}
    \label{tab:AE_EH_archetypal_clustering_F1F2}
  \end{center}
    
\end{table}

\begin{table}[t]
  \begin{center}
    \small\addtolength{\tabcolsep}{-1pt}
	\begin{tabular}{|c|c|c|}
    \hline
    \multicolumn{3}{|c|}{Singaporean Children} \\
    \hline
   	\multirow{2}{*}{Phoneme} & Group1 & Group2 \\
     &  (767, 2183) & (726, 1978) \\
   	\hline
   	/\textipa{\ae}/ & \textbf{0.594} & 0.406  \\
   	/\textipa{E}/ & \textbf{0.604}  & 0.396 \\
   	\hline
   	\multicolumn{3}{c}{} \\
   	\hline
   	\multicolumn{3}{|c|}{American Children} \\
   	\hline
   	\multirow{2}{*}{Phoneme} & Group1 & Group2 \\
   	 & (872, 2188) &  (816, 1957) \\
   	\hline
   	 /\textipa{\ae}/ & \textbf{0.550} & 0.450 \\
   	 /\textipa{E}/ & 0.436 & \textbf{0.564} \\
   	\hline
   	\multicolumn{3}{c}{} \\
   	\hline
   	\multicolumn{3}{|c|}{British Children} \\
   	\hline
   	 \multirow{2}{*}{Phoneme} & Group1 & Group2 \\
   	  &  (891, 1656) & (795, 1972) \\
   	\hline
   	/\textipa{\ae}/ &  \textbf{0.976}  & 0.024 \\
   	/\textipa{E}/ & 0.167 &  \textbf{0.833} \\
   	\hline
    \end{tabular}
    \caption{Kmeans clustering using \textit{F1}(Hz), \textit{F2}(Hz) estimates of /\textipa{\ae}/ and /\textipa{E}/from Singaporean, American and British children.  Centroids for each cluster are in the format (F1, F2).}
    \label{tab:AE_EH_kmeans_clustering_F1F2}
  \end{center}
    
\end{table}

\subsection{Acoustic Analysis and Characterization}
\label{subsec:ae_eh_acoustic}

\subsubsection{F1 F2 formant space}
\begin{figure}
	\centering
	\includegraphics[width=\linewidth]{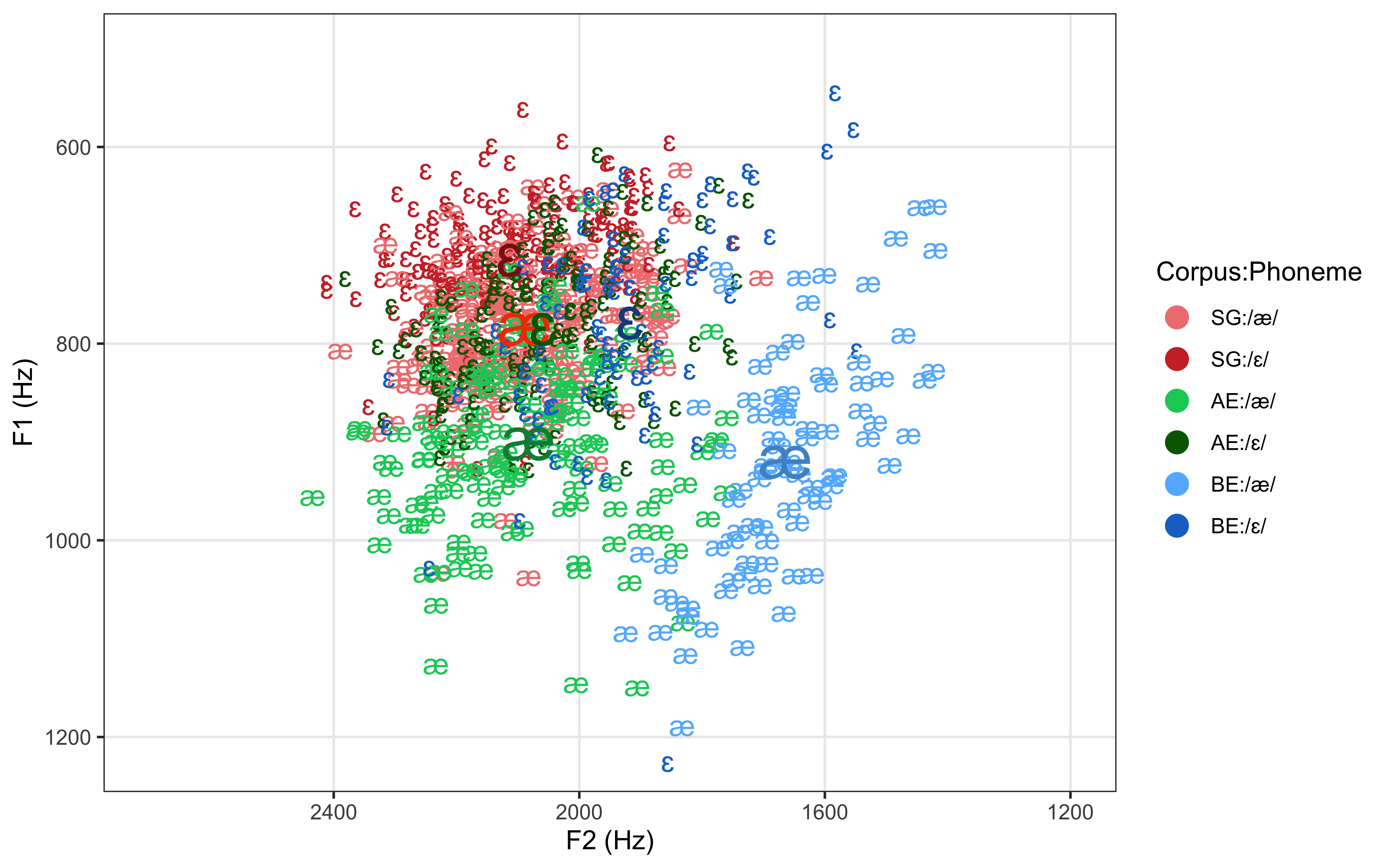}
    \caption{Singaporean and American children are more similar in terms of formant characteristics for /\textipa{\ae}/ and /\textipa{E}/ vowels and they show less distinction between these vowels in comparison to British children. 
    Smaller points: individual speaker's mean \textit{F1} and \textit{F2}; larger points: group means for the speaker groups.}
    \label{fig:AE_EH_comparison}
\end{figure}

Estimates of F1 and F2 for  /\textipa{\ae}/ and /\textipa{E}/ in the three speaker groups are summarized in Table \ref{tab:AE_EH_overall_table}. Using each speaker as a data point, we visualize this differences between speaker groups for the two vowels in Figure \ref{fig:AE_EH_comparison}. We observe that  /\textipa{\ae}/ and /\textipa{E}/ produced by British children are the most clearly distinguished from each other, with /\textipa{E}/ having a higher F2 and lower F1 than that of /\textipa{\ae}/. However, the F1 and F2 distinction between these two vowels for Singaporean and American children are less conspicuous. These observations in the formant space align with our clustering results, potentially explaining why the split for Singaporean and American children's /\textipa{\ae}/ and /\textipa{E}/ are not as clear as that of British children. We then further analyzed differences across the speaker groups in terms of F1 and F2 separately.

\begin{table}[t]
  \begin{center}
  \small\addtolength{\tabcolsep}{-1pt}
  \begin{tabular}[\linewidth]{|c|c|c|c|c|c|}
	\hline
	\bf Corpus & Phoneme & F1 mean & F1 se & F2 mean & F2 se \\
	\hline
      SG & /\textipa{\ae}/ & 785 &  4.83 & 2089 & 8.71 \\
      & /\textipa{E}/ & 715 &  4.14 & 2113 & 8.95 \\ 
      \hline
      AE & /\textipa{\ae}/ & 902 & 7.38 & 2083 &  12.4 \\
      & /\textipa{E}/ & 786 & 5.88 & 2060 & 11.6 \\
      \hline
      BE & /\textipa{\ae}/ & 921 & 12.7 & 1665 & 12.8 \\
      & /\textipa{E}/ & 779 & 11.8 & 1918 & 19.6 \\
	\hline
  \end{tabular}
  \caption{Mean and standard error (se) for each speaker group for /\textipa{\ae}/ and /\textipa{E}/ formant estimates.}
  \label{tab:AE_EH_overall_table}
  \end{center}
 \end{table}

\subsubsection{F1 formant estimates} 
For /\textipa{\ae}/, Singaporean children have the lowest F1 (\textit{M} = 785), American children have higher F1 estimates (\textit{M} = 902), and British children have the highest F1 (\textit{M} = 921). Similarly, for the production of /\textipa{E}/, Singaporean children have the lowest F1 (\textit{M} = 715), American children have higher F1 estimates (\textit{M} = 786), which is similar to British children (\textit{M} = 779). One-way ANOVA tests demonstrated that for both vowels, these differences are statistically significant, \textit{F}(2, 412) = 108.7, \textit{p} $<$ 0.001 for /\textipa{\ae}/ and \textit{F}(2, 413) = 43.26, \textit{p} $<$ 0.001 for /\textipa{E}/. We followed up our ANOVA tests with post hoc Tukey's HSD Test which show that Singaporean children's average F1 for /\textipa{\ae}/ is significantly lower than that of American children (\textit{p} $<$ 0.001) and British children (\textit{p} $<$ 0.001). However, the difference between American and British speakers is not significant (\textit{p} $=$ 0.244). Similarly, Singaporean children's average F1 for /\textipa{E}/ is significantly lower than those of British (\textit{p} $<$ 0.001) children and American children (\textit{p} $<$ 0.001), but there is no significant difference between the latter two (\textit{p} $=$ 0.801). In terms of articulatory implication, this suggests that compared to American and British children, Singaporean children produce both /\textipa{\ae}/ and /\textipa{E}/ vowels with a higher tongue position.



\subsubsection{F2 formant estimates} 
For /\textipa{\ae}/, Singaporean children (\textit{M} = 2089) and American children (\textit{M} = 2083) have similarly high F2, while British children have a much lower F2 (\textit{M} = 1665). The same trend is observed for /\textipa{E}/, where Singaporean children also have the highest F2 (\textit{M} = 2113), American children have slightly lower F2 (\textit{M} = 2060), and British children have the lowest F2 (\textit{M} = 1918). One-way ANOVA showed that for both vowels, these differences are statistically significant, \textit{F}(2, 412) = 353.8, \textit{p} $<$ 0.001 for /\textipa{\ae}/ and \textit{F}(2, 413) = 55.79, \textit{p} $<$ 0.001 for /\textipa{E}/. We followed up our ANOVA tests with post hoc Tukey's HSD Test, which show that for both /\textipa{\ae}/ and /\textipa{E}/, Singaporean and American children's F2 are significantly higher (\textit{p} $<$ 0.001) than those of British children. Within Singaporean and American children, they are not significantly different (\textit{p} $=$ 0.916) in terms of F2 for /\textipa{\ae}/. For /\textipa{E}/, Singaporean children demonstrate higher F2 compared to American children (\textit{p} $<$ 0.01).  In terms of articulatory implications, Singaporean and American children consistently exhibit fronting of both /\textipa{\ae}/ and /\textipa{E}/ such that they are produced with a much fronter tongue position compared to those of British children. Within Singaporean and American children, their /\textipa{\ae}/ vowels are produced with similar frontness while for /\textipa{E}/, Singaporean children demonstrate even more fronting than American children.

\subsection{Summary: Singaporean and American children similarly demonstrate fronting of /\textipa{\ae}/ and /\textipa{E}/ vowels and show less clear distinction between these vowels compared to British children}
\label{subsec:ae_eh_overall}


Inferring from the acoustic characterizations of /\textipa{\ae}/ and /\textipa{E}/, the corresponding articulatory implication is that Singaporean children, compared to American and British children, are producing these vowels with a higher tongue position. Interestingly, this tendency for producing vowels with a higher tongue position, is not only observed for /\textipa{\ae}/ and /\textipa{E}/ but also for the TRAP$-$BATH split [\textipa{\ae}] and [\textipa{A}] vowels we examined earlier in subsection \ref{subsec:trapbath_acoustic}.  Compared to British children, Singaporean and American children are more similar in terms of their fronting of both /\textipa{\ae}/ and /\textipa{E}/. This fronting behavior also bears resemblance to the trend we observed for TRAP$-$BATH split [\textipa{\ae}] and [\textipa{A}] vowels in subsection \ref{subsec:trapbath_acoustic}. With the fronting of both /\textipa{\ae}/ and /\textipa{E}/, Singaporean and American children do not show as much distinction between these two vowels compared to British children as the two vowels are produced with similar frontness that is more characteristic of /\textipa{E}/ than /\textipa{\ae}/.

These trends observed from formant analysis agree with and provide further explanations for the results from our clustering experiments. Particularly, the fronting of /\textipa{\ae}/ and /\textipa{E}/ by Singaporean and American children such that these two vowels are produced with similar frontness (of the tongue position) likely explains why  /\textipa{\ae}/ and /\textipa{E}/ from these two speaker populations do not fall into clean distinct clusters like those of British speakers in our unsupervised clustering experiments.

\section{/\textipa{A}/ vs. /\textipa{O}/}
\label{sec:aa_ao}
/\textipa{A}/ and /\textipa{O}/ are both back vowels, with /\textipa{O}/ having slightly lower F1 and F2 estimates (difference in F2 estimates even less prominent) compared to the /\textipa{A}/ phoneme \citep{Stevens:98}. \citet{Ladefoged:15} noted that many Midwestern and Californian speakers of American English do not distinguish [\textipa{A}] and [\textipa{O}] as in \textit{cot} and \textit{caught} and usually have a vowel intermediate, whereas many speakers of British English have an additional vowel in this area and distinguish between the vowels [\textipa{A}], [\textipa{6}], [\textipa{O}] as  \textit{balm}, \textit{bomb} and \textit{bought}.

Given the acoustic characteristics of the two phonemes, producing /\textipa{A}/ with a higher tongue position, resulting in lower F1, could lead to potential confusion with /\textipa{O}/. Having observed that Singaporean children exhibit lower F1 in their TRAP$-$BATH split vowels (across realizations as [\textipa{\ae}] and [\textipa{A}]  phones) in subsection \ref{subsec:trapbath_acoustic} as well as in /\textipa{\ae}/ and /\textipa{E}/ vowels in subsection \ref{subsec:ae_eh_acoustic}, we further examine if they show similar traits for /\textipa{A}/ and /\textipa{O}/ vowels. We also investigate how production of /\textipa{A}/ and /\textipa{O}/ phonemes might differ across the three speaker groups.

\subsection{Unsupervised clustering}
\label{subsec:aa_ao_unsup}
We perform per-speaker-group clustering across /\textipa{A}/ and /\textipa{O}/ vowels from each group  using F1 and F2 estimates. Clustering results using Archetypal analysis and Kmeans clustering are shown in Tables \ref{tab:AA_AO_archetypal_clustering_F1F2} and \ref{tab:AA_AO_kmeans_clustering_F1F2}. The results from the two clustering methods both show that /\textipa{A}/ and /\textipa{O}/ vowels produced by each of the speaker groups are largely (about at least 80 \%) grouped into two clean, distinctive clusters. This distinction is only less clear for British children's /\textipa{A}/ vowels when using Archetypal analysis (67.1 \% in majority group).

\begin{table}[t]
  \begin{center}
    \small\addtolength{\tabcolsep}{-1pt}
	\begin{tabular}{|c|c|c|}
    \hline
    \multicolumn{3}{|c|}{Singaporean Children} \\
    \hline
   	\multirow{2}{*}{Phoneme} & Group1 & Group2 \\
     &  (664, 1015) & (1043, 1847) \\
   	\hline
   	/\textipa{A}/ & 0.193 & \textbf{0.807} \\
   	/\textipa{O}/  & \textbf{0.974} & 0.026 \\
   	\hline
   	\multicolumn{3}{c}{} \\
   	\hline
   	\multicolumn{3}{|c|}{American Children} \\
   	\hline
   	\multirow{2}{*}{Phoneme} & Group1 & Group2 \\
     & (593, 1032) &  (1073, 1623) \\
   	\hline
   	/\textipa{A}/ & 0.157 & \textbf{0.843} \\
   	/\textipa{O}/  & \textbf{0.907} & 0.093 \\
   	\hline
   	\multicolumn{3}{c}{} \\
   	\hline
   	\multicolumn{3}{|c|}{British Children} \\
   	\hline
   	 \multirow{2}{*}{Phoneme} & Group1 & Group2 \\
     &  (474, 1020) & (841, 1442) \\
   	\hline
   	/\textipa{A}/ & 0.329 & \textbf{0.671} \\
   	/\textipa{O}/  & \textbf{0.904} & 0.096 \\
   	\hline
    \end{tabular}
    \caption{Archetypal analysis using \textit{F1}(Hz), \textit{F2}(Hz) estimates of /\textipa{A}/ and /\textipa{O}/ from Singaporean, American and British children.  Archetypal extreme points for each cluster are in the format (F1, F2).}
    \label{tab:AA_AO_archetypal_clustering_F1F2}
  \end{center}
    
\end{table}

\begin{table}[t]
  \begin{center}
    \small\addtolength{\tabcolsep}{-1pt}
	\begin{tabular}{|c|c|c|}
    \hline
    \multicolumn{3}{|c|}{Singaporean Children} \\
    \hline
   	\multirow{2}{*}{Phoneme} & Group1 & Group2 \\
     &  (751, 1187) & (899, 1550) \\
   	\hline
   	/\textipa{A}/ & 0.052 & \textbf{0.948} \\
   	/\textipa{O}/  & \textbf{0.969} & 0.031 \\
   	\hline
   	\multicolumn{3}{c}{} \\
   	\hline
   	\multicolumn{3}{|c|}{American Children} \\
   	\hline
   	\multirow{2}{*}{Phoneme} & Group1 & Group2 \\
     &  (764, 1236) & (935, 1461) \\
   	\hline
   	/\textipa{A}/ & 0.200 & \textbf{0.800} \\
   	/\textipa{O}/  & \textbf{0.936} & 0.064 \\
   	\hline
   	\multicolumn{3}{c}{} \\
   	\hline
   	\multicolumn{3}{|c|}{British Children} \\
   	\hline
   	 \multirow{2}{*}{Phoneme} & Group1 & Group2 \\
     &  (562, 1126) & (697, 1271) \\
   	\hline
   	/\textipa{A}/ & 0.129 & \textbf{0.871} \\
   	/\textipa{O}/  & \textbf{0.795} & 0.205 \\
   	\hline
    \end{tabular}
    \caption{Kmeans clustering using \textit{F1}(Hz), \textit{F2}(Hz) estimates of /\textipa{A}/ and /\textipa{O}/ from Singaporean, American and British children.  Centroids for each cluster are in the format (F1, F2).}
    \label{tab:AA_AO_kmeans_clustering_F1F2}
  \end{center}
    
\end{table}

\subsection{Acoustic Analysis and Characterization}
\label{subsec:aa_ao_acoustic}
 \subsubsection{F1 F2 formant space} 
\begin{figure}
	\centering
	\includegraphics[width=1.16\linewidth]{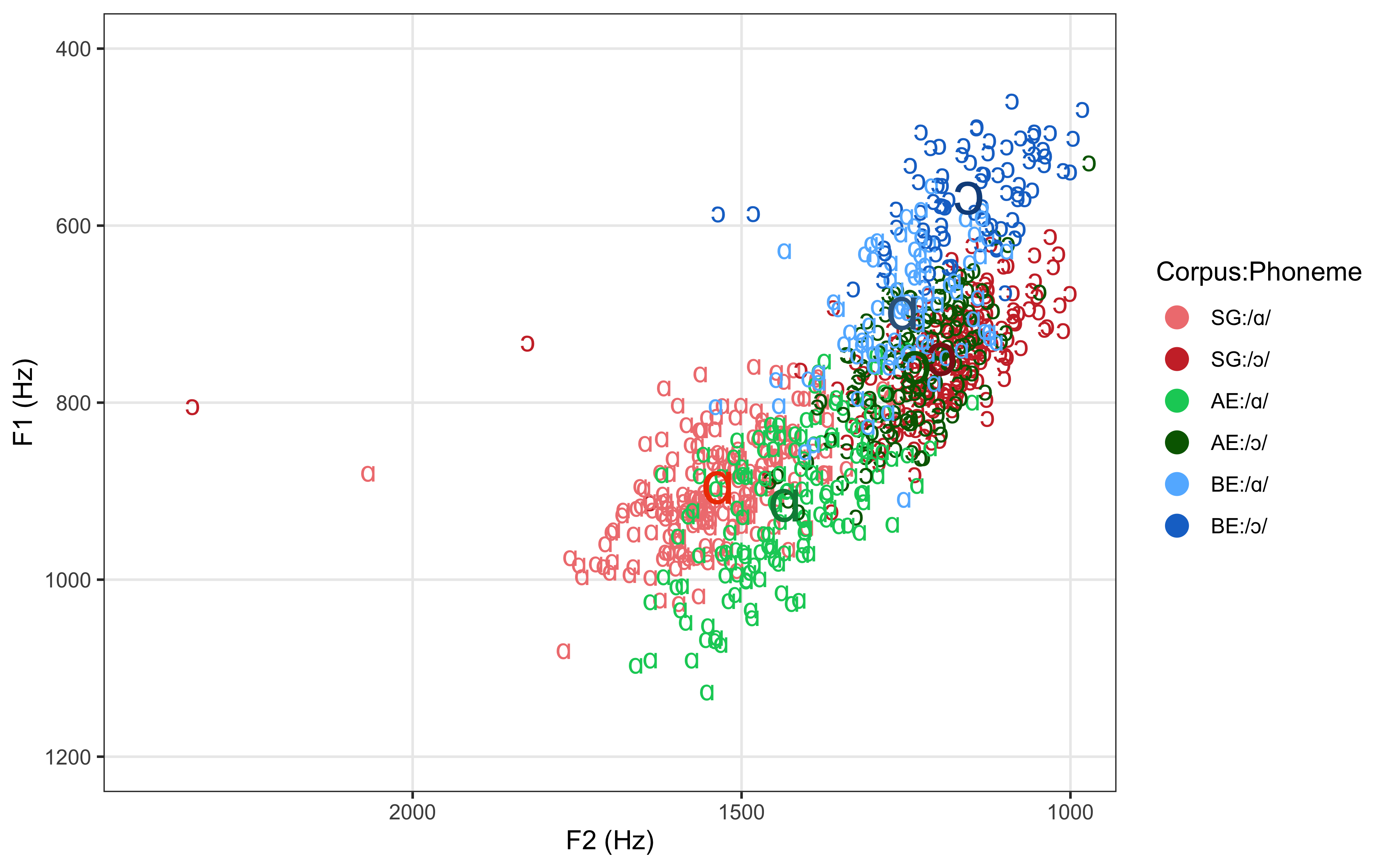}
    \caption{Singaporean and American children are more similar in  terms of \textit{F1} and \textit{F2} estimates for /\textipa{A}/ and /\textipa{O}/ vowels compared to British children. Smaller points: individual speaker's mean \textit{F1} and \textit{F2}; larger points: group means for the speaker groups.}
    \label{fig:AA_AO_comparison}
\end{figure}

 We present the mean and standard error for F1 and F2 estimates in Table \ref{tab:AA_AO_table} and visualize this data on a per speaker level in Figure \ref{fig:AA_AO_comparison}. We observe that for all three speaker groups, the /\textipa{A}/ and /\textipa{O}/ vowels are largely distinguished from each other, with /\textipa{O}/ vowels being produced with lower F1 and F2 estimates. 
The /\textipa{A}/ vowels produced by Singaporean children share similar F1 and F2 characteristics as those produced by American children. This similarity is also observed for the /\textipa{O}/ vowels produced by the two speaker groups. Compared to the /\textipa{A}/ and /\textipa{O}/ vowels produced by Singaporean and American children, the two vowels seem closer in the formant space and show slight overlapping for British speakers. We further analyzed these formant characteristics in detail in terms of F1 and F2 separately.

\begin{table}[t]
  \begin{center}
  \small\addtolength{\tabcolsep}{-1pt}
  \begin{tabular}[\linewidth]{|c|c|c|c|c|c|}
	\hline
	\bf Corpus & Phone & F1 mean & F1 se & F2 mean & F2 se \\
	\hline
      SG & [\textipa{A}] & 896 & 4.43  & 1535 &  7.40 \\
      & [\textipa{O}] & 751 &  4.15 &  1195 & 9.13 \\ 
      \hline
      AE & [\textipa{A}] & 916 &  6.98 & 1433 & 8.45 \\
      & [\textipa{O}] & 760 &  5.83 & 1234 & 6.59 \\ 
      \hline
      BE & [\textipa{A}] & 699 &  7.63 &  1255 & 9.32 \\
      & [\textipa{O}] & 569 &  5.79 & 1153 & 10.4 \\ 
	\hline
  \end{tabular}
  \caption{Mean and standard error (se) for each speaker group for /\textipa{A}/ and /\textipa{O}/ formant estimates.}
  \label{tab:AA_AO_table}
  \end{center}
\end{table}

\subsubsection{F1 formant estimates} 
For /\textipa{A}/, British children have the lowest F1 (\textit{M} = 699), Singaporean children have higher F1 estimates (\textit{M} = 896), and American children show slightly higher F1 (\textit{M} = 916) compared to that of Singaporean children. Similarly, for the production of /\textipa{O}/, British children have the lowest F1 (\textit{M} = 569), and the F1 estimates for Singaporean children (\textit{M} = 751) and American children (\textit{M} = 760) are close to each other and higher than that of British children. One-way ANOVA tests demonstrated that for both vowels, these differences are statistically significant, \textit{F}(2, 414) = 286.5, \textit{p} $<$ 0.001 for /\textipa{A}/ and \textit{F}(2, 412) = 311.9, \textit{p} $<$ 0.001 for /\textipa{O}/. 
We followed up our ANOVA tests with post hoc Tukey's HSD Test which show that British children's average F1 for /\textipa{A}/ and /\textipa{O}/ vowels are significantly lower than those of Singaporean and American children (\textit{p} $<$ 0.001). However, the difference between Singaporean and American speakers is not significant (\textit{p} $=$ 0.0284 for /\textipa{A}/ and \textit{p} $=$ 0.364 for /\textipa{O}/). In terms of articulatory implication, this suggests that compared to Singaporean and American children, British children produce both /\textipa{A}/ and /\textipa{O}/ vowels with a higher tongue position.

\subsubsection{F2 formant estimates}
For /\textipa{A}/, Singaporean children (\textit{M} = 1535) and American children (\textit{M} = 1433) both have higher F2 compared to that of British children (\textit{M} = 1255). Whereas for /\textipa{O}/, the F2 estimates across the three speaker groups are more similar, with American children having the highest F2 (\textit{M} = 1234), Singaporean children having just slightly lower F2 (\textit{M} = 1195), and British children having the lowest F2 (\textit{M} = 1153). One-way ANOVA showed that for both vowels, these differences are statistically significant, \textit{F}(2, 414) = 239.7, \textit{p} $<$ 0.001 for /\textipa{A}/ and \textit{F}(2, 412) = 15.28, \textit{p} $<$ 0.001 for /\textipa{O}/. 
We followed up our ANOVA tests with post hoc Tukey's HSD Test, which show that for both /\textipa{A}/ and /\textipa{O}/, Singaporean and American children's F2 estimates are significantly higher (\textit{p} $<$ 0.01) than those of British children. This suggests that in terms of articulation, Singaporean and American children exhibit fronting of both /\textipa{A}/ and /\textipa{O}/ such that they are produced with a more fronted tongue position compared to those of British children. Between Singaporean and American children, Singaporean children's F2 estimates are higher for /\textipa{A}/ (\textit{p} $<$ 0.001) and American children's are higher for /\textipa{O}/ (\textit{p} $=$ 0.00299).

\subsection{Summary: Singaporean and American children similarly demonstrate lower but more fronted tongue positions for /\textipa{A}/ and /\textipa{O}/ vowels compared to British children and show clear distinction between the two vowels}
\label{subsec:aa_ao_overall}
Compared to British children, Singaporean and American children are more similar in producing their /\textipa{A}/ and /\textipa{O}/ vowels with a lower tongue position and in fronting both vowels. The observation that British children are producing these vowels with a higher tongue position (lower F1) such that their /\textipa{A}/ and /\textipa{O}/ vowels are produced with more similar heights possibly explains the unsupervised clustering results that distinction between the two vowels is only less clear for British children (when using Archetypal analysis). 

F1 characteristics of /\textipa{A}/ and /\textipa{O}/ revealed an interesting trend for this vowel pair where British children articulate these vowels with higher vowel heights, in contrast to the trends we observed for TRAP$-$BATH split, /\textipa{\ae}/, and /\textipa{E}/ vowels (subsections \ref{subsec:trapbath_acoustic}, \ref{subsec:ae_eh_acoustic}) where it is usually Singaporean and sometimes also American children who produce the vowels with a higher tongue position. In terms of the frontness of the vowels, Singaporean and American children's fronting of /\textipa{A}/ and /\textipa{O}/ is similar to the trends we observed for TRAP$-$BATH split [\textipa{\ae}] and [\textipa{A}] vowels in subsection \ref{subsec:trapbath_acoustic}, as well as for /\textipa{\ae}/ and /\textipa{E}/ in subsection \ref{subsec:ae_eh_acoustic}. 


\section{Tense and lax vowels - /\textipa{u}/ vs. /\textipa{U}/}
\label{sec:sec_long_short_u}
We can characterize the differences between tense and lax vowels by considering them in pairs. In each tense and lax vowel pair, the lax vowel is produced with a shorter duration, lower tongue position and is slightly more centralized in the formant space compared to the corresponding tense vowel \citep{Ladefoged:15}. One such pair is [\textipa{u}, \textipa{U}] as in \textit{boot}, \textit{foot}, where /\textipa{u}/ is a tense vowel and /\textipa{U}/ is a lax vowel. In terms of acoustic characteristics, this means that the /\textipa{U}/ vowel would have higher F1 and F2 estimates compared to the tense vowel /\textipa{u}/.\footnote{\citep{Ladefoged:15} also points that there can be variations amongst American and British speakers} Further, the duration of /\textipa{U}/ would also be shorter than that of /\textipa{u}/.

Given these characteristics, if /\textipa{u}/ is produced with slightly higher F1 and F2 estimates, /\textipa{U}/ is produced with slightly lower F1 and F2 estimates, or the duration difference is less conspicuous, the two vowels can be easily confused. We observed in sections \ref{sec:trap_bath}, \ref{sec:ae_eh} and \ref{sec:aa_ao} that some speaker groups tend to produce some vowels with a higher tongue position, or show fronting of some vowels. In this section, we investigate if the speaker groups show similar trends for the /\textipa{u}/ and /\textipa{U}/ vowel pair, and how that might affect the distinction between the tense and lax vowels in this pair.

\subsection{Unsupervised clustering}
\label{subsec:sec_long_short_u_unsup}
We perform per-speaker-group clustering across /\textipa{u}/ and /\textipa{U}/ vowels from each speaker group, using F1, F2 and duration estimates as features. The results from unsupervised clustering experiments using Archetypal analysis and Kmeans clustering are summarized in Tables \ref{tab:long_short_u_archetypal_clustering_F1F2} and \ref{tab:long_short_u_kmeans_clustering_F1F2}. The Archetypal analysis results indicate that /\textipa{u}/ and /\textipa{U}/ vowels produced by Singaporean children are similar as they largely fall into the same group, whereas /\textipa{u}/ and /\textipa{U}/ vowels produced by American and British children largely ($\geq$ 90 \%) fall into two clean, distinctive groups. For this vowel pair, Kmeans clustering generally give less clean groups, and a clear distinction ($\geq$ 90 \%)  between the two vowels is only observed for British children.

\begin{table}[t]
  \begin{center}
    \small\addtolength{\tabcolsep}{-1pt}
	\begin{tabular}{|c|c|c|}
    \hline
    \multicolumn{3}{|c|}{Singaporean Children} \\
    \hline
   	\multirow{2}{*}{Phoneme} & Group1 & Group2 \\
     & (474, 1212, 0.180) &  (702, 1616, 0.0726)  \\
   	\hline
   	/\textipa{u}/ & \textbf{0.927} & 0.073  \\
   	/\textipa{U}/ & \textbf{0.698} & 0.302 \\
   	\hline
   	\multicolumn{3}{c}{} \\
   	\hline
   	\multicolumn{3}{|c|}{American Children} \\
   	\hline
   	\multirow{2}{*}{Phoneme} & Group1 & Group2 \\
    & (656, 1691, 0.0667) &  (401, 1623, 0.221) \\
   	\hline
   	/\textipa{u}/ & 0.100 & \textbf{0.900} \\
   	/\textipa{U}/ & \textbf{1.000} & 0.000  \\
   	\hline
   	\multicolumn{3}{c}{} \\
   	\hline
   	\multicolumn{3}{|c|}{British Children} \\
   	\hline
   	 \multirow{2}{*}{Phoneme} & Group1 & Group2 \\
     &  (558, 1603, 0.0594) & (395, 2230, 0.150) \\
   	\hline
   	/\textipa{u}/ & 0.000 & \textbf{1.000} \\
   	/\textipa{U}/  & \textbf{0.988} & 0.012 \\
   	\hline
    \end{tabular}
    \caption{Archetypal analysis using \textit{F1}(Hz), \textit{F2}(Hz), duration(seconds) estimates of /\textipa{u}/ and /\textipa{U}/ from Singaporean, American and British children. Archetypal extreme points for each cluster are in the format (F1, F2, duration).}
    \label{tab:long_short_u_archetypal_clustering_F1F2}
  \end{center}
    
\end{table}

\begin{table}[t]
  \begin{center}
    \small\addtolength{\tabcolsep}{-1pt}
	\begin{tabular}{|c|c|c|}
    \hline
    \multicolumn{3}{|c|}{Singaporean Children} \\
    \hline
   	\multirow{2}{*}{Phoneme} & Group1 & Group2 \\
     &  (545, 1223, 0.146) & (551, 1500, 0.134) \\
   	\hline
   	/\textipa{u}/ & 0.458 & \textbf{0.542} \\
   	/\textipa{U}/  & \textbf{0.724} & 0.276 \\
   	\hline
   	\multicolumn{3}{c}{} \\
   	\hline
   	\multicolumn{3}{|c|}{American Children} \\
   	\hline
   	\multirow{2}{*}{Phoneme} & Group1 & Group2 \\
    & (546, 1562, 0.134)  &  (536, 1797, 0.135) \\
   	\hline
   	/\textipa{u}/ & 0.493 & \textbf{0.507} \\
   	/\textipa{U}/  & \textbf{0.664} & 0.336 \\
   	\hline
   	\multicolumn{3}{c}{} \\
   	\hline
   	\multicolumn{3}{|c|}{British Children} \\
   	\hline
   	 \multirow{2}{*}{Phoneme} & Group1 & Group2 \\
     &  (522, 1726, 0.0840) & (430, 2106, 0.124) \\
   	\hline
   	/\textipa{u}/ & 0.098 & \textbf{0.902} \\
   	/\textipa{U}/  & \textbf{ 0.976} & 0.024 \\
   	\hline
    \end{tabular}
    \caption{Kmeans clustering using \textit{F1}(Hz), \textit{F2}(Hz), duration(seconds) estimates of /\textipa{u}/ and /\textipa{U}/ from Singaporean, American and British children. Centroids for each cluster are in the format (F1, F2, duration).}
    \label{tab:long_short_u_kmeans_clustering_F1F2}
  \end{center}
    
\end{table}

\subsection{Acoustic and Duration Analysis}
\label{subsec:sec_long_short_u_acoustic}

\subsubsection{F1 F2 formant space}
\begin{figure}
	\centering
	\includegraphics[width=\linewidth]{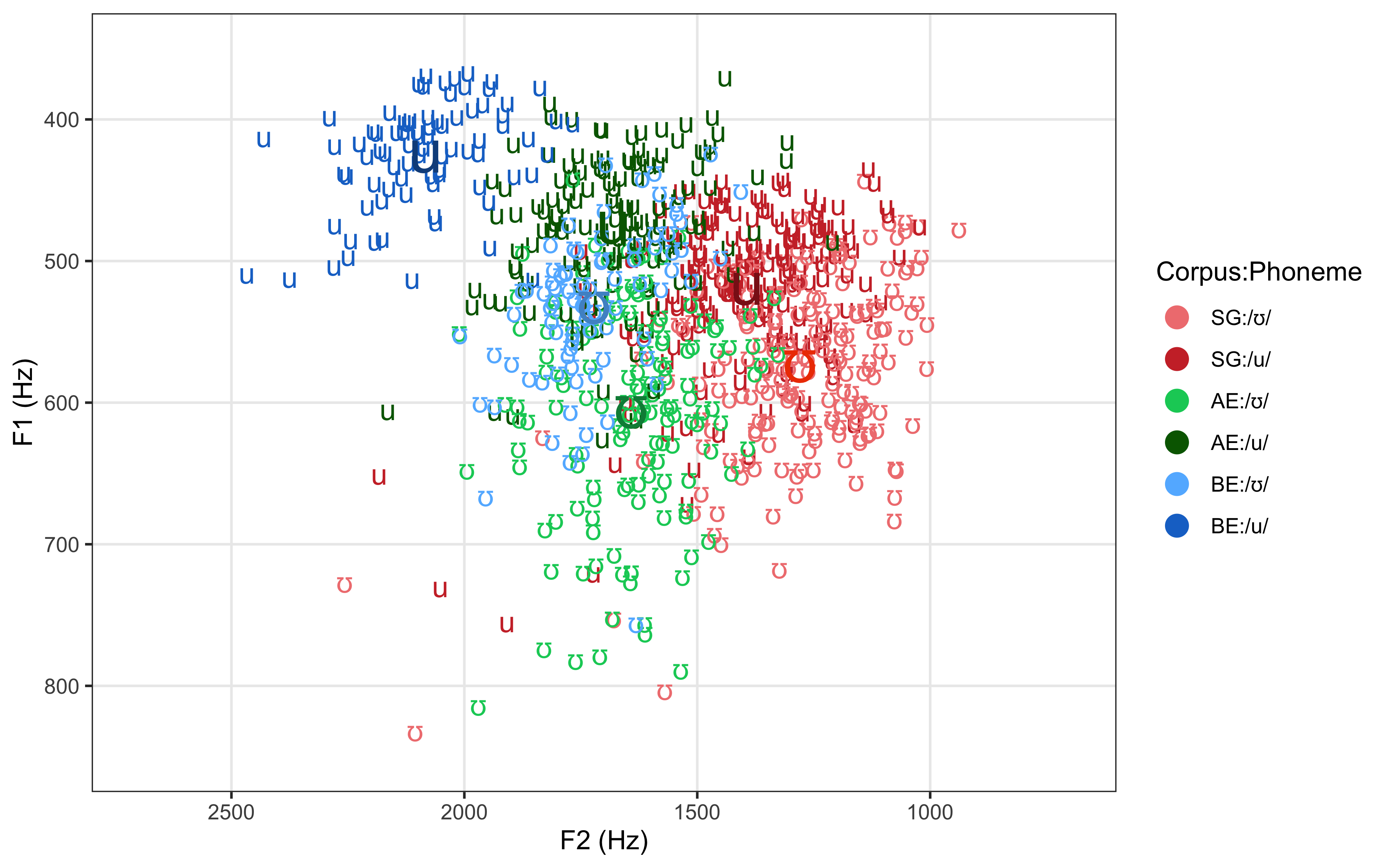}
    \caption{Singaporean children are more similar to American children in  terms of \textit{F1} and \textit{F2} for /\textipa{u}/ and /\textipa{U}/ vowels. Smaller points: individual speaker's mean \textit{F1} and \textit{F2}; larger points: group means for the speaker groups.}
    \label{fig:UH_UW_formants_comparison}
\end{figure}

Estimates of F1 and F2 for /\textipa{u}/ and /\textipa{U}/ in the three speaker groups are summarized in Table \ref{tab:long_short_u_overall_table}. Using each speaker as a data point, we visualize the differences between speaker groups for these two vowels in Figure \ref{fig:UH_UW_formants_comparison}. We observe that /\textipa{u}/ and /\textipa{U}/ produced by British children are the most clearly distinguished from each other, with /\textipa{U}/ having higher F1 and lower F2 estimates than that of /\textipa{u}/. However, the F1 and F2 distinction between these two vowels for Singaporean and American children are less conspicuous -- \textipa{u}/ and /\textipa{U}/ share relatively more similar F2 estimates for American children, and the two vowel groups are generally close to each other in formant space for Singaporean children. We further analyzed this tense and lax vowel pair produced by the speaker groups in detail, looking into F1 and F2 characteristics separately, and also in terms of vowel duration.


\begin{table}[t]
  \begin{center}
  \small\addtolength{\tabcolsep}{-1pt}
  \begin{tabular}[\linewidth]{|c|c|c|c|c|c|}
	\hline
	\bf Corpus & Phoneme & F1 mean & F1 se & F2 mean & F2 se \\
	\hline
      SG & /\textipa{u}/ & 520 & 3.86 & 1393 & 11.9 \\
      & /\textipa{U}/ & 574 &  4.54 & 1279 & 12.7 \\ 
      \hline
      AE & /\textipa{u}/ & 477 &  4.28 & 1680 & 12.8 \\
      & /\textipa{U}/ & 607 &  6.26 & 1641 & 11.7 \\
      \hline
      BE & /\textipa{u}/ & 426 & 4.15 & 2083 & 15.4 \\
      & /\textipa{U}/ & 533 &  6.27 & 1722 & 13.7 \\
	\hline
  \end{tabular}
  \caption{Mean and standard error (se) for each speaker group for /\textipa{u}/ and /\textipa{U}/ formant estimates.}
  \label{tab:long_short_u_overall_table}
  \end{center}
 \end{table}

\subsubsection{F1 formant estimates}
For /\textipa{u}/, British children have the lowest F1 (\textit{M} = 426), America children have higher F1 estimates (\textit{M} = 477), and Singaporean children show the highest F1 (\textit{M} = 520) amongst the three speaker groups. For the production of /\textipa{U}/, it is similar in that British children also have the lowest F1 (\textit{M} = 533), and the F1 estimates for Singaporean children (\textit{M} = 574) and American children (\textit{M} = 607) are higher than that of British children. One-way ANOVA tests demonstrated that for both vowels, these differences are statistically significant, \textit{F}(2, 411) = 105.4, \textit{p} $<$ 0.001 for /\textipa{u}/ and \textit{F}(2, 411) = 32.89, \textit{p} $<$ 0.001 for /\textipa{U}/.  
We followed up our ANOVA tests with post hoc Tukey's HSD Test which show that British children's average F1 for /\textipa{u}/ and /\textipa{U}/ vowels are significantly lower than those of Singaporean and American children (\textit{p} $<$ 0.001). In terms of articulatory implication, this suggests that compared to Singaporean and American children, British children produce both /\textipa{u}/ and /\textipa{U}/ vowels with a higher tongue position. Between Singaporean and American children, Singaporean children's tongue positions are lower for /\textipa{u}/ (\textit{p} $<$ 0.001) and American children's are lower for /\textipa{U}/ (\textit{p} $<$ 0.001).



\subsubsection{F2 formant estimates} 
For /\textipa{u}/, Singaporean children have the lowest F2 estimiates (\textit{M} = 1393), American children have higher F2 estimates (\textit{M} = 1680), and British children have the highest F2 (\textit{M} = 2083). The same trend is observed for /\textipa{U}/, where Singaporean children also have the lowest F2 (\textit{M} = 1279), American children have higher F2 (\textit{M} = 1641) and British children demonstrate the highest F2(\textit{M} = 1722). One-way ANOVA showed that for both vowels, these differences are statistically significant, \textit{F}(2, 411) = 578.3, \textit{p} $<$ 0.001 for /\textipa{u}/ and \textit{F}(2, 411) = 336.8, \textit{p} $<$ 0.001 for /\textipa{U}/. We followed up our ANOVA tests with post hoc Tukey's HSD Test, which show that for both /\textipa{u}/ and /\textipa{U}/, Singaporean and American children's F2 estimates are significantly lower (\textit{p} $<$ 0.001) than those of British children. This suggests that in terms of articulation, Singaporean and American children's /\textipa{u}/ and /\textipa{U}/ are produced with a less fronted tongue position compared to those of British children. Between Singaporean and American children, it is consistent across the two vowels that Singaporean children demonstrate even less fronted tongue position compared to those of American children (\textit{p} $<$ 0.001).

\subsubsection{Duration}
\begin{figure}
	\centering
	\includegraphics[width=\linewidth]{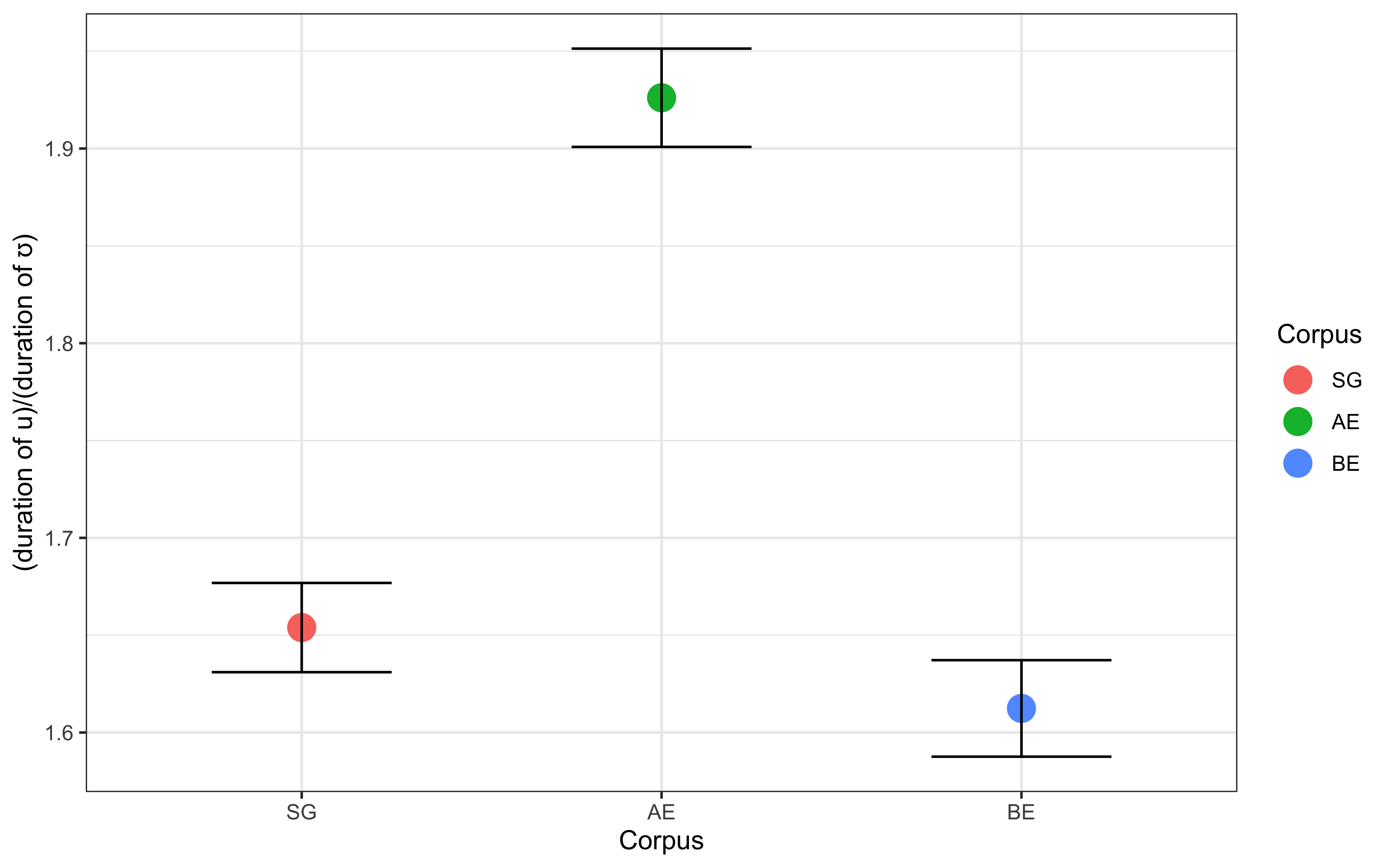}
    \caption{/\textipa{u}/ and /\textipa{U}/ duration ratio comparison across speaker groups. All groups show clear duration distinction between the two vowels, with /\textipa{u}/ being produced with a vowel duration least 1.6 times that of /\textipa{U}/. Colored points mark the mean duration ratio for each speaker group, whereas the error bars indicate standard error.}
    \label{fig:UH_UW_duration_comparison}
\end{figure}

We compare the differences in vowel duration between the tense and lax vowels for /\textipa{u}/ and /\textipa{U}/. T-tests for correlated samples revealed that all three speaker groups produce /\textipa{u}/ vowels that are of significantly longer duration compared to their /\textipa{U}/ vowels (Singaporean children: $t(191) =$ 33.12, \textit{p} $<$ 0.001 (two-tailed); American children: $t(139) =$ 36.09, \textit{p} $<$ 0.001 (two-tailed); British children: $t(81) =$ 23.64, \textit{p} $<$ 0.001 (two-tailed)).  We then investigate if this duration distinction is even more prominent for some speaker group(s). 
Figure \ref{fig:UH_UW_duration_comparison} shows the trend for duration ratio across speaker groups where for each speaker, the /\textipa{u}/ and /\textipa{U}/ duration ratio is calculated by 
\[
\text{duration ratio} = \frac{\text{average duration of /\textipa{u}/ from that speaker}}{\text{average duration of /\textipa{U}/ from that speaker}}.
\]
One-way ANOVA showed that the differences in duration ratio across speaker groups are statistically significant, \textit{F}(2, 411) = 43.71, \textit{p} $<$ 0.001. Post hoc Tukey's HSD Test show that the /\textipa{u}/ and /\textipa{U}/ duration ratio for Singaporean (\textit{M} = 1.65) and British children (\textit{M} = 1.61) are significantly smaller (\textit{p} $<$ 0.001) compared to American children (\textit{M} = 1.93). However, the /\textipa{u}/ and /\textipa{U}/ duration ratio for Singaporean and British children are not significantly different (\textit{p} $=$ 0.535).

\subsection{Summary: Singaporean and American children similarly demonstrate lower and less fronted tongue positions for /\textipa{u}/ and /\textipa{U}/ vowels compared to British children. All groups show clear duration distinction between the two vowels.}
\label{subsec:sec_long_short_u_overall}
Singaporean children's /\textipa{u}/ and /\textipa{U}/ vowels are more similar to American children in that they are produced with a lower tongue position (higher F1 estimates) compared to British children. This trend is similarly observed when the speakers produce other back vowels like /\textipa{A}/ and /\textipa{O}/ (subsection \ref{subsec:aa_ao_acoustic}). Whereas, for TRAP$-$BATH split, /\textipa{\ae}/, and /\textipa{E}/ vowels (subsections \ref{subsec:trapbath_acoustic}, \ref{subsec:ae_eh_acoustic}), it was Singaporean and sometimes also American children who usually produce the vowels with a higher tongue position than that of British children. 

Singaporean children's /\textipa{u}/ and /\textipa{U}/ are also similar to American children in that they are produced with a less fronted tongue position (lower F2 estimates) compared to those of British children. Interestingly, this characteristic of Singaporean children's /\textipa{u}/ and /\textipa{U}/ tense lax vowel pair is different from when they produce other vowels like the TRAP$-$BATH split [\textipa{\ae}] and [\textipa{A}] vowels, /\textipa{\ae}/,  /\textipa{E}/, /\textipa{A}/ and /\textipa{O}/ (subsections \ref{subsec:trapbath_acoustic}, \ref{subsec:ae_eh_acoustic}, \ref{subsec:aa_ao_acoustic}) where they consistently demonstrated fronting of the vowels.

In terms of F1 and F2 characteristics, as seen in Figure \ref{fig:UH_UW_formants_comparison}, British children show the greatest distinction with /\textipa{u}/ being produced with a higher tongue height and more fronted tongue position than /\textipa{U}/. This possibly explains the results from Kmeans clustering where clear distinction ($\geq$ 90 \%)  between the two vowels is only observed for British children. Although all speakers show significant duration difference between /\textipa{u}/ and /\textipa{U}/, the difference is the most prominent for American children. Compared to British and American children, Singaporean children neither mostly clearly distinguishes /\textipa{u}/ and /\textipa{U}/ through formant feature differences nor duration ratio differences. This potentially accounts for why the /\textipa{u}/ and /\textipa{U}/ tokens produced by Singaporean children do not fall into two distinct groups for our unsupervised clustering experiments, a trend that is consistently observed across Archetypal analysis and Kmeans clustering.

Finally, it is interesting that across children in all three speaker groups, /\textipa{U}/ is articulated with a lower F2 (if not, similar) compared to /\textipa{u}/, which differs from the usual canonical characteristics of /\textipa{u}/ and /\textipa{U}/. \citep{Ladefoged:15}'s chapter on ``English Vowels" also illustrated diagrams showing how there could be such variations amongst American and British speakers. From our work, such variation may also be an interesting characteristic in children speech and worth further investigation in future work.

\section{Tense and lax vowels  - /\textipa{i}/ vs. /\textipa{I}/}
\label{sec:sec_long_short_i}
Another tense and lax vowel pair is [\textipa{i}, \textipa{I}] as in \textit{beat}, \textit{bit}. The lax vowel /\textipa{I}/ is shorter, lower, and more centralized compared to the corresponding tense vowel /\textipa{i}/ \citep{Ladefoged:15}. In terms of acoustic characteristics, this means that the /\textipa{I}/ vowel would have higher F1 and lower F2 estimates compared to the tense vowel /\textipa{i}/. Further, the duration of /\textipa{I}/ would also be shorter than that of /\textipa{i}/.

Given these characteristics, if /\textipa{i}/ is produced with slightly higher F1 and lower F2 estimates, /\textipa{I}/ is produced with slightly lower F1 and higher F2 estimates, or the duration difference is less conspicuous, the two vowels can be easily confused. It is consistent across the previous sections (sections \ref{sec:trap_bath}, \ref{sec:ae_eh}, \ref{sec:aa_ao}, \ref{sec:sec_long_short_u}) that some speaker groups tend to produce particular vowels with a higher tongue position, or show fronting of particular vowels. We also observed that the effect may differ based on the vowel concerned. In this section, we investigate the trends for the /\textipa{i}/ and /\textipa{I}/ vowel pair, and make comparisons across different speaker groups.

\subsection{Unsupervised clustering}
\label{subsec:sec_long_short_i_unsup}
We perform per-speaker-group clustering across /\textipa{i}/ and /\textipa{I}/ vowels from each speaker group using F1, F2 and vowel duration estimates. Clustering results using Archetypal analysis and Kmeans clustering are shown in Tables \ref{tab:long_short_i_archetypal_clustering_F1F2} and \ref{tab:long_short_i_kmeans_clustering_F1F2}. The Archetypal analysis results indicate that  /\textipa{i}/ and /\textipa{I}/ vowels produced by American and British children largely ($>$ 90 \%) fall into two clean, distinctive groups, whereas the distinction is less clear for the /\textipa{i}/ and /\textipa{I}/ produced by Singaporean children. Kmeans clustering generally give less clean groups, but the results similarly show that the distinction between the two vowels for Singaporean children is less clear compared to the other two speaker groups.

\begin{table}[t]
  \begin{center}
    \small\addtolength{\tabcolsep}{-1pt}
	\begin{tabular}{|c|c|c|}
    \hline
    \multicolumn{3}{|c|}{Singaporean Children} \\
    \hline
   	\multirow{2}{*}{Phoneme} & Group1 & Group2 \\
     &  (507, 2241, 0.0902) & (526, 2899, 0.279) \\
   	\hline
   	/\textipa{i}/ & 0.344  & \textbf{0.656}  \\
   	/\textipa{I}/  & \textbf{0.844} & 0.156 \\
   	\hline
   	\multicolumn{3}{c}{} \\
   	\hline
   	\multicolumn{3}{|c|}{American Children} \\
   	\hline
   	\multirow{2}{*}{Phoneme} & Group1 & Group2 \\
    &  (607, 2142, 0.0801) & (363, 2898, 0.242)  \\
   	\hline
   	/\textipa{i}/ & 0.086  & \textbf{0.914} \\
   	/\textipa{I}/  & \textbf{0.993} & 0.007 \\
   	\hline
   	\multicolumn{3}{c}{} \\
   	\hline
   	\multicolumn{3}{|c|}{British Children} \\
   	\hline
   	 \multirow{2}{*}{Phoneme} & Group1 & Group2 \\
      & (564, 2040, 0.0622) &  (403, 2779, 0.232)\\
   	\hline
   	/\textipa{i}/ & 0.071 & \textbf{0.929} \\
   	/\textipa{I}/  & \textbf{1.000} & 0.000 \\
   	\hline
    \end{tabular}
    \caption{Archetypal analysis using \textit{F1}(Hz), \textit{F2}(Hz), duration(seconds) estimates of /\textipa{i}/ and /\textipa{I}/ from Singaporean, American and British children. Archetypal extreme points for each cluster are in the format (F1, F2, duration).}
    \label{tab:long_short_i_archetypal_clustering_F1F2}
  \end{center}
    
\end{table}

\begin{table}[t]
  \begin{center}
    \small\addtolength{\tabcolsep}{-1pt}
	\begin{tabular}{|c|c|c|}
    \hline
    \multicolumn{3}{|c|}{Singaporean Children} \\
    \hline
   	\multirow{2}{*}{Phoneme} & Group1 & Group2 \\
     & (513, 2370, 0.148) &  (517, 2658, 0.193) \\
   	\hline
   	/\textipa{i}/ & 0.266 & \textbf{0.734} \\
   	/\textipa{I}/  & \textbf{0.583} & 0.417  \\
   	\hline
   	\multicolumn{3}{c}{} \\
   	\hline
   	\multicolumn{3}{|c|}{American Children} \\
   	\hline
   	\multirow{2}{*}{Phoneme} & Group1 & Group2 \\
    & (547, 2288, 0.125)  &  (437, 2735, 0.184) \\
   	\hline
   	/\textipa{i}/ & 0.271 & \textbf{0.729} \\
   	/\textipa{I}/  & \textbf{0.971} & 0.029  \\
   	\hline
   	\multicolumn{3}{c}{} \\
   	\hline
   	\multicolumn{3}{|c|}{British Children} \\
   	\hline
   	 \multirow{2}{*}{Phoneme} & Group1 & Group2 \\
     &  (515, 2179, 0.107) & (460, 2598, 0.169) \\
   	\hline
   	/\textipa{i}/ & 0.274 & \textbf{0.726} \\
   	/\textipa{I}/  & \textbf{0.881} & 0.119 \\
   	\hline
    \end{tabular}
    \caption{Kmeans clustering using \textit{F1}(Hz), \textit{F2}(Hz), duration(seconds) estimates of /\textipa{i}/ and /\textipa{I}/ from Singaporean, American and British children. Centroids for each cluster are in the format (F1, F2, duration).}
    \label{tab:long_short_i_kmeans_clustering_F1F2}
  \end{center}
    
\end{table}

\subsection{Acoustic and Duration Analysis}
\label{subsec:sec_long_short_i_acoustic}

\subsubsection{F1 F2 formant space}
\begin{figure}
	\centering
	\includegraphics[width=\linewidth]{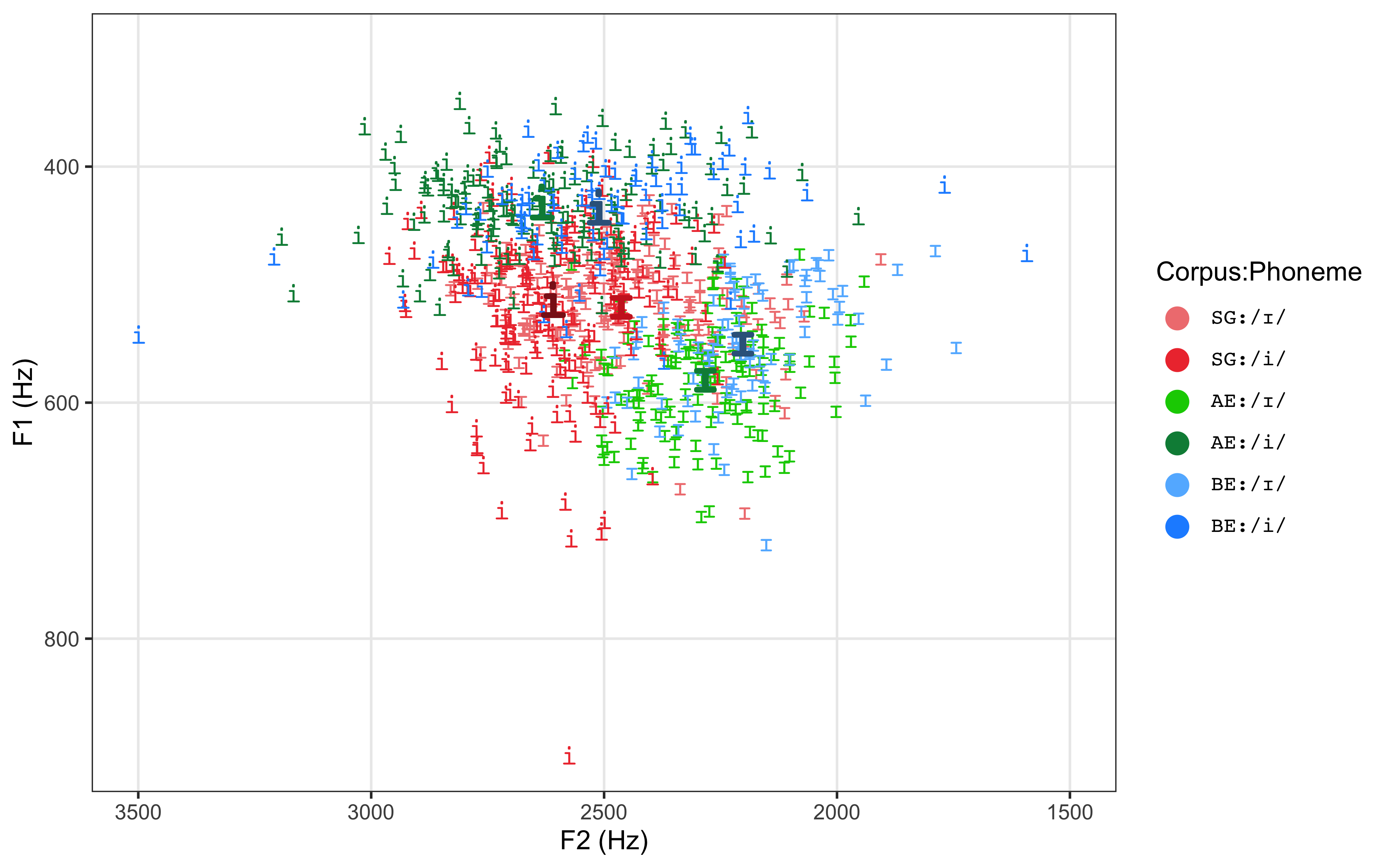}
    \caption{Compared to American and British children, Singaporean children's /\textipa{i}/ and /\textipa{I}/ are less distinguished in terms of \textit{F1} and \textit{F2} estimates. Smaller points: individual speaker's mean \textit{F1} and \textit{F2}; larger points: group means for the speaker groups.}
    \label{fig:IY_IH_formants_comparison}
\end{figure}

Estimates of F1 and F2 for /\textipa{i}/ and /\textipa{I}/ in the three speaker groups are summarized in Table \ref{tab:long_short_i_overall_table}. Using each speaker as a data point, we visualize the differences between speaker groups for the two vowels in Figure \ref{fig:IY_IH_formants_comparison}. We observe that the /\textipa{i}/ and /\textipa{I}/ vowels produced by Singaporean children are close to each other in the formant space. This observation is different from both that for American and British children whose  /\textipa{i}/ and /\textipa{I}/ vowels form more distinctively separate groups in the formant space, with /\textipa{I}/ having a higher F1 and lower F2 than that of /\textipa{i}/. 


\begin{table}[t]
  \begin{center}
  \small\addtolength{\tabcolsep}{-1pt}
  \begin{tabular}[\linewidth]{|c|c|c|c|c|c|}
	\hline
	\bf Corpus & Phoneme & F1 mean & F1 se & F2 mean & F2 se \\
	\hline
      SG & /\textipa{i}/ & 515 &  4.99 &  2609 &  10.5 \\
      & /\textipa{I}/ & 516 & 2.96 & 2463 & 12.7 \\ 
      \hline
      AE & /\textipa{i}/ & 432 & 3.09 & 2633 & 19.1 \\
      & /\textipa{I}/ & 578 & 3.78 & 2282 & 11.7 \\
      \hline
      BE & /\textipa{i}/ & 437 &  4.55 & 2511 & 28.3 \\
      & /\textipa{I}/ & 547 & 5.27 & 2201 & 17.0 \\
	\hline
  \end{tabular}
  \caption{Mean and standard error (se) for each speaker group for /\textipa{i}/ and /\textipa{I}/ formant estimates.}
  \label{tab:long_short_i_overall_table}
  \end{center}
 \end{table}

\subsubsection{F1 formant estimates}
Singaporean children's F1 estimates for /\textipa{i}/ (\textit{M} = 515) is much higher compared to that of American (\textit{M} = 432) and British children (\textit{M} = 437). Whereas, for the production of /\textipa{I}/, Singaporean children have the lowest F1 (\textit{M} = 516), the F1 estimates for British children is higher (\textit{M} = 547) and American children (\textit{M} = 578) show the highest F1 estimates amongst the three groups. One-way ANOVA tests demonstrated that for both vowels, these differences are statistically significant, \textit{F}(2, 413) = 112.8, \textit{p} $<$ 0.001 for /\textipa{i}/ and \textit{F}(2, 413) = 81.24, \textit{p} $<$ 0.001 for /\textipa{I}/.  We followed up our ANOVA tests with post hoc Tukey's HSD Test which show that Singaporean children's F1 estimates for /\textipa{i}/ are significantly higher (\textit{p} $<$ 0.001) than those of American and British children, while American and British children are not significantly different from each other  (\textit{p} $=$ 0.832). For /\textipa{I}/, American children's F1 estimates are significantly higher than that of the other two groups (\textit{p} $<$ 0.001). Between Singaporean and British children, Singaporean children's F1 estimates for /\textipa{I}/  are even lower than British children.  In terms of articulatory implication, this suggests that amongst the three speaker groups, Singaporean children produce /\textipa{i}/ with the lower tongue position and /\textipa{I}/ with a higher tongue position compared to the other two speaker groups. American children produce /\textipa{I}/ with the lowest tongue position.



\subsubsection{F2 formant estimates}
For /\textipa{i}/, British children have the lowest F2 estimates (\textit{M} = 2511), Singaporean (\textit{M} = 2609) and American children (\textit{M} = 2633) have higher F2 estimates which are more similar to each other. For /\textipa{I}/, British children also have the lowest F2 (\textit{M} = 2201), American children have higher F2 (\textit{M} = 2282) and Singapore children have the highest F2 (\textit{M} = 2463). One-way ANOVA showed that for both vowels, these differences are statistically significant, \textit{F}(2, 413) = 10.2, \textit{p} $<$ 0.001 for /\textipa{i}/ and \textit{F}(2, 413) = 96.38, \textit{p} $<$ 0.001 for /\textipa{I}/. We followed up our ANOVA tests with post hoc Tukey's HSD Test, which show that Singaporean and American children's F2 estimates for /\textipa{i}/ are significantly higher (\textit{p} $<$ 0.001) than those of British children, whereas the difference within Singaporean and American children is not significant (\textit{p} $=$ 0.535). For /\textipa{I}/, all pairwise differences are significant (\textit{p} $<$ 0.001). Singaporean and American children's F2 estimates for /\textipa{I}/ are also significantly higher (\textit{p} $<$ 0.001) than those of British children. Between Singaporean and American children, Singaporean children's F2 estimates for /\textipa{I}/ are even higher (\textit{p} $<$ 0.001) than American children, indicating that they demonstrate more fronting. These suggest that in terms of articulation, Singaporean and American children's /\textipa{i}/ and /\textipa{I}/ are produced with a more fronted tongue position compared to those of British children.

\subsubsection{Duration}
\begin{figure}[t]
	\centering
	\includegraphics[width=\linewidth]{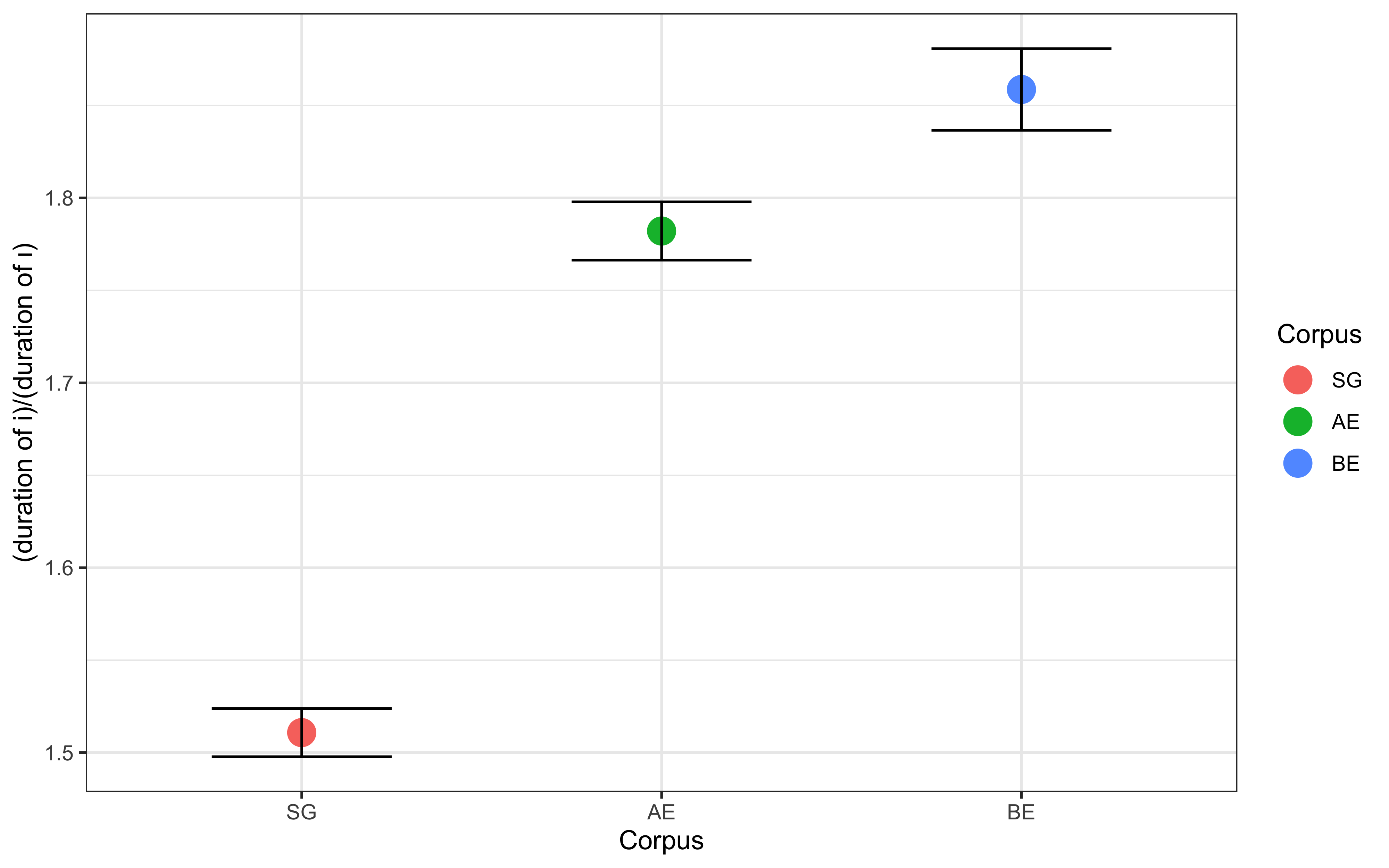}
    \caption{/\textipa{i}/ and /\textipa{I}/ duration comparison across speaker groups. All groups show clear duration distinction between the two vowels, with /\textipa{i}/ being produced with a vowel duration least 1.5 times that of /\textipa{I}/. Colored points mark the mean duration ratio for each speaker group, whereas the error bars indicate standard error.}
    \label{fig:IY_IH_duration_comparison}
\end{figure}

We compare the differences in vowel duration between the tense and lax vowel pair for /\textipa{i}/ and /\textipa{I}/. T-tests for correlated samples revealed that all three speaker groups the produce /\textipa{i}/ vowels that are of significantly longer duration compared to their /\textipa{I}/ vowels (Singaporean children: $t(191) =$ 39.28, \textit{p} $<$ 0.001 (two-tailed); American children: $t(139) =$ 47.12, \textit{p} $<$ 0.001 (two-tailed); British children: $t(83) =$ 21.83, \textit{p} $<$ 0.001 (two-tailed)).  We then investigate if this duration distinction is more prominent for some speaker group(s). 
Figure \ref{fig:IY_IH_duration_comparison} shows the trend for duration ratio across speaker groups where for each speaker, the /\textipa{i}/ and /\textipa{I}/ duration ratio is calculated by
\[
\text{duration ratio} = \frac{\text{average duration of /\textipa{i}/ from that speaker}}{\text{average duration of /\textipa{I}/ from that speaker}}.
\]
One-way ANOVA showed that the differences in duration ratio across speaker groups are statistically significant, \textit{F}(2, 413) = 137.2, \textit{p} $<$ 0.001. Post hoc Tukey's HSD Test show that the /\textipa{i}/ and /\textipa{I}/ duration ratio for Singaporean children (\textit{M} = 1.51) is significantly smaller (\textit{p} $<$ 0.001) compared to those of American (\textit{M} = 1.78) and British children (\textit{M} = 1.86). Between American and British children, the duration ratio is even greater for British children (\textit{p} $=$ 0.00902). Therefore, the characteristic of /\textipa{i}/ being produced with a longer vowel duration compared to /\textipa{I}/ is significantly more prominent in American and British children compared to Singaporean children, and is most prominent in British children.

\subsection{Summary: Singaporean children produce /\textipa{i}/ with a lower tongue position and /\textipa{I}/ with a higher tongue position compared to the other two speaker groups. They show the least distinction between this vowel pair.}
\label{subsec:sec_long_short_i_overall}
Amongst the three groups, Singaporean children produce the tense vowel /\textipa{i}/ with the lowest tongue position and American children produce the lax vowel /\textipa{I}/ with the lowest tongue position (highest F1 estimates). This trend bears resemblance to what we have observed for the tense lax vowel pair /\textipa{u}/ and /\textipa{U}/ (subsection \ref{subsec:sec_long_short_u_acoustic}), where Singaporean children also produce the tense vowel /\textipa{u}/  with the lowest tongue position and American children produce the lax vowel /\textipa{U}/ with the lowest tongue position amongst the speaker groups.


Singaporean children's /\textipa{i}/ and /\textipa{I}/ are more similar to those of American children in that they are produced with a more fronted tongue position compared to those of British children. This aligns with the trends we observed when these speakers produce other vowels like the TRAP$-$BATH split [\textipa{\ae}] and [\textipa{A}] vowels, /\textipa{\ae}/, /\textipa{E}/, /\textipa{A}/, and  /\textipa{O}/ (sections \ref{subsec:trapbath_acoustic}, \ref{subsec:ae_eh_acoustic}, \ref{subsec:aa_ao_acoustic}) where they consistently demonstrated fronting of the vowels compared to British children.

In terms of F1 and F2 characteristics, as seen in Figure \ref{fig:IY_IH_formants_comparison}, American and British children show greater distinction between this tense lax vowel pair with /\textipa{i}/ being produced with a higher tongue height and more fronted tongue position than /\textipa{I}/. Singaporean children tend to produce /\textipa{i}/ with a lower tongue position and /\textipa{I}/ with a higher tongue position compared to the other groups such that their /\textipa{i}/ and /\textipa{I}/ are not significantly different in terms of vowel height, $t(191) =$ 0.311, \textit{p} $=$ 0.756 (two-tailed). Further, Singaporean children also consistently exhibit fronting of the vowels such that the vowels are closer in terms of the frontness. For duration differences, although all speaker groups show significant duration differences between /\textipa{i}/ and /\textipa{I}/, the difference is the more prominent for British and American children. These trends support why /\textipa{i}/ and /\textipa{I}/ for Singaporean children give less clean clusters in the unsupervised clustering experiments. Therefore, exploratory analysis using supervised learning, formant space analysis and vowel duration analysis all provide evidence that point towards Singaporean children distinguishing less clearly between /\textipa{i}/ and /\textipa{I}/ in their articulation compared to American and British children.


\section{Analysis of lateral approximant /\textipa{l}/}
The lateral approximant /\textipa{l}/ has traditionally been observed to have two variants, namely, dark /\textipa{l}/ which appears in syllable rimes (syllable-final /\textipa{l}/) and clear /\textipa{l}/ in syllable onsets (syllable initial /\textipa{l}/) \citep{sweet1923sounds, Johnes1947phonetics}. Previous studies such as \citet{Hansen2001LinguisticCO} and \citet{He2014ProductionOE} have shown that speakers of Mandarin experienced difficulties in producing syllable-final /\textipa{l}/, and tend to modify syllable-final /\textipa{l}/ in their production. Given that many Singaporeans are bilingual speakers of both English and Mandarin, and the Chinese languages spoken \cite{deterding2007ethnic} in Singapore, we devote this section towards investigating if this sociolinguistic background affects Singaporean children's production of syllable-final /\textipa{l}/ like the way Mandarin speakers' production are affected \citep{Hansen2001LinguisticCO,He2014ProductionOE}.

Acoustic characteristics of syllable-final /\textipa{l}/ include 1) lowering of F1 and F2 formant features when transitioning from a preceding vowel \cite{Johnson:11}, 2) a relatively wide gap between F2 and F3 \cite{Liberman:88, Stevens:98}, and 3) F1 and F2 are expected to have values close to each other when the phoneme is present \cite{Bayley:15}. We focus on these formant features and compare how the production of syllable-final /\textipa{l}/ vary across Singaporean, American and British children.

\subsection{Unsupervised clustering}

Archetypal analysis and Kmeans clustering on syllable-final /\textipa{l}/ tokens across the three speakers groups were performed using the first three formants as features. The results for Archetypal analysis are summarized in Table \ref{tab:dark_l_archetypal_clustering_F1F2F3} whereas the results obtained using Kmeans clustering are summarized in Table \ref{tab:dark_l_kmeans_clustering_F1F2F3}. For each row, the two numbers add up to 1.0 (stands for 100\%). This illustrates the proportion of each speaker group's tokens that gets grouped into Group 1 and Group 2 respectively. Both sets of results show that syllable-final /\textipa{l}/ tokens produced by Singaporean children are mostly grouped into a different group from those produced by British children speakers, whereas those produced by American children is somewhere between those two groups with more tokens being grouped into the British dominant group. To quantify this difference, detailed acoustic analysis using the characteristic formant features of syllable-final /\textipa{l}/ tokens was carried out.

\begin{table}[t]
  \begin{center}
    \small\addtolength{\tabcolsep}{-1pt}
	\begin{tabular}{|c|c|c|}
    \hline
   	\multirow{2}{*}{Corpus} & Group1 & Group2 \\
    &  (830, 1870, 3450) & (419, 1040, 3001)  \\
   	\hline
   	SG & \textbf{0.651} & 0.349 \\
   	AE & 0.414 & \textbf{0.586} \\
   	BE  & 0.012 & \textbf{0.988} \\
   	\hline
    \end{tabular}
    \caption{Archetypal analysis using \textit{F1}(Hz), \textit{F2}(Hz), and \textit{F3}(Hz) estimates of syllable-final  /\textipa{l}/ from Singaporean, American and British children. Archetypal extreme points for each cluster are in the format (F1, F2, F3).}
    \label{tab:dark_l_archetypal_clustering_F1F2F3}
  \end{center}
    
\end{table}

\begin{table}[t]
  \begin{center}
    \small\addtolength{\tabcolsep}{-1pt}
	\begin{tabular}{|c|c|c|}
     \hline
   	\multirow{2}{*}{Corpus} & Group1 & Group2 \\
    &  (661, 1558, 3298) & (554, 1278, 3114)  \\
   	\hline
   	SG & \textbf{0.755} & 0.245 \\
   	AE & 0.471 & \textbf{0.529} \\
   	BE  & 0.061 & \textbf{0.939} \\
   	\hline
    \end{tabular}
    \caption{Kmeans clustering using \textit{F1}(Hz), \textit{F2}(Hz), and \textit{F3}(Hz) estimates of syllable-final  /\textipa{l}/ from Singaporean, American and British children. Centroids for each cluster are in the format (F1, F2, F3).}
    \label{tab:dark_l_kmeans_clustering_F1F2F3}
  \end{center}
    
\end{table}

\subsection{Acoustic Analysis}
Our acoustic analysis serves to compare how well the syllable-final /\textipa{l}/ tokens produced by each speaker group demonstrates the standard acoustic features of the phoneme as reported in literature, in terms of the first three formants.

\subsubsection{Falling F1} 
When transitioning into the syllable-final /\textipa{l}/ phoneme, we anticipate a lowering of F1 as compared to that in the preceding vowel \cite{Johnson:11}. That is, if we calculate the F1 difference between the syllable-final /\textipa{l}/ and the preceding vowel using Equation \ref{eq:F1Diff}, we expect a negative value when a characteristic syllable-final /\textipa{l}/ has been produced, or in some cases, due to the formant features of the preceding vowel, a very small value.

\begin{equation}
    \begin{split} \label{eq:F1Diff}
        \text{F1 difference} = \text{Average F1 in syllable-final /\textipa{l}/} - \\ \text{Average F1 in preceding vowel}
    \end{split}
\end{equation}

\begin{figure}[t]
	\centering
	\includegraphics[width=\linewidth]{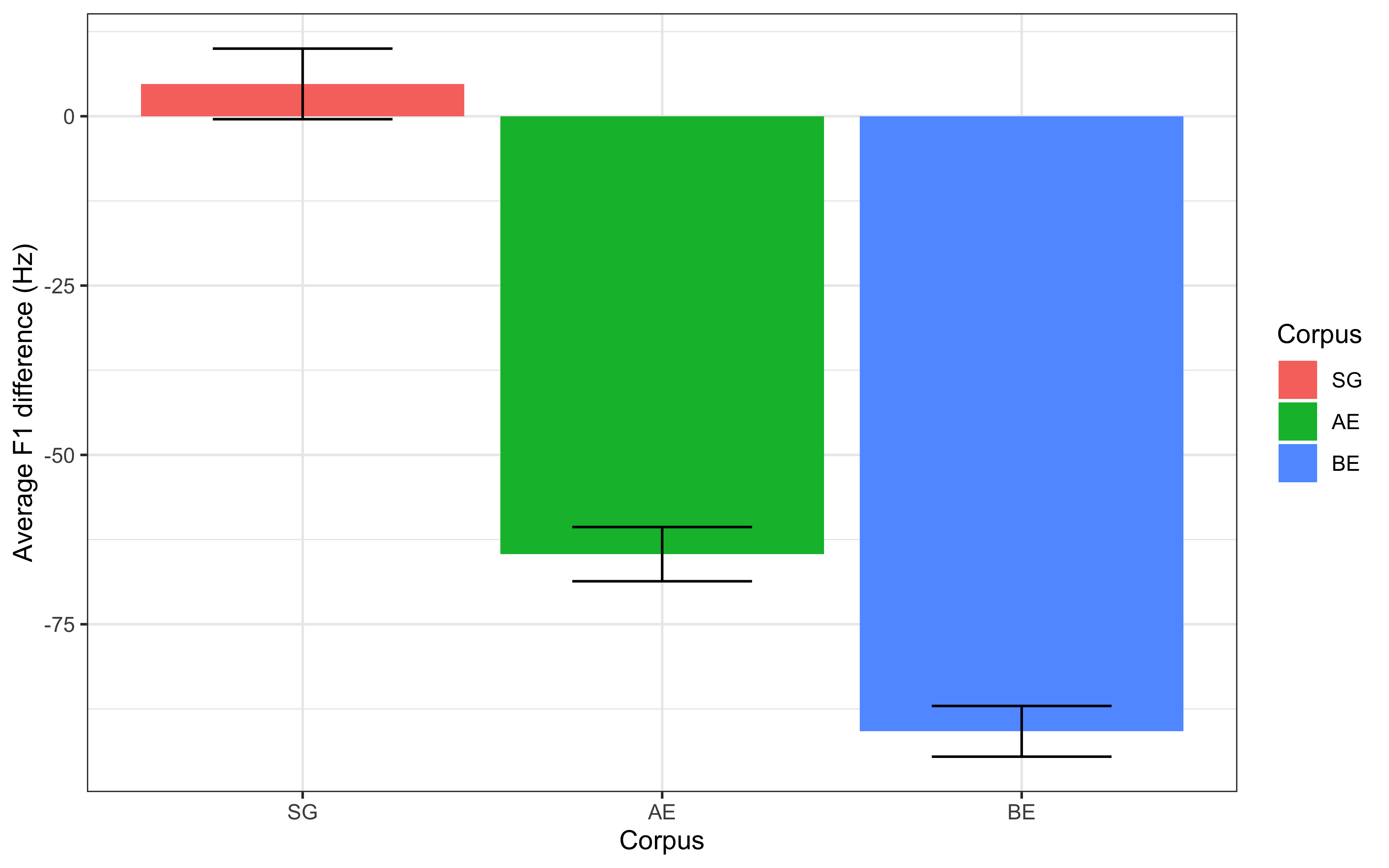}
    \caption{Comparison of F1 transition into syllable-final /\textipa{l}/. Singaporean children do not show conspicuous lowering of F1, compared to the other two speaker groups. The F1 difference plotted is calculated using the formula in Equation \ref{eq:F1Diff} on a per speaker basis. The error bars indicate standard error.}
    \label{fig:darkL_general_f1transition_comparison}
\end{figure}

From the visualization of F1 differences during transition as shown in Figure \ref{fig:darkL_general_f1transition_comparison}, syllable-final /\textipa{l}/ tokens produced by American (\textit{M} = -64.7) and British children (\textit{M} = -90.8) show a decrease in F1 estimates whereas Singaporean children demonstrated a very slight increase (\textit{M} = 4.77). One-way ANOVA demonstrated that these differences are statistically significant across speaker groups, \textit{F}(2, 825) = 99.72, \textit{p} $<$ 0.001.  We then followed up our ANOVA test with a post hoc Tukey's HSD Test which showed that, significant difference was observed between Singaporean and American children speakers, as well as Singaporean and British children speakers at \textit{p} $<$ 0.001 but not between American and British children speakers (\textit{p} $=$ 0.00377). This shows that, where a syllable-final /\textipa{l}/ phoneme is involved, Singaporean children speakers maybe producing something that is acoustically much less characteristic of a typical syllable-final /\textipa{l}/ compared to other speakers in terms of F1 characteristics. 

\begin{figure}[t]
	\centering
	\includegraphics[width=\linewidth]{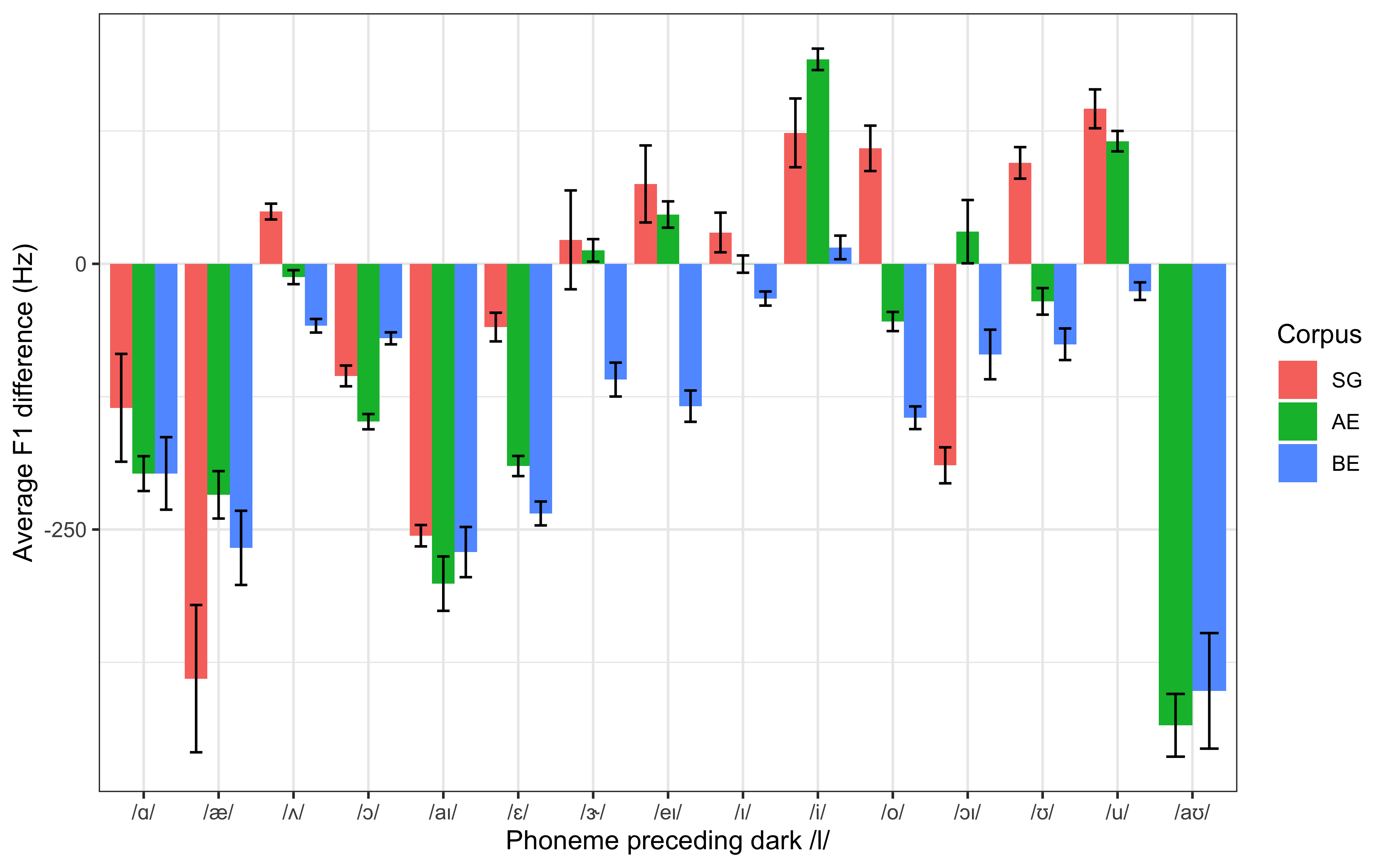}
    \caption{Syllable final /\textipa{l}/ F1 transition comparison by preceding phoneme across speaker groups. Compared to the other two speaker groups, Singaporean children generally do not demonstrate lowering of F1 as clearly in this transition. The error bars indicate standard error. (Syllable final /\textipa{l}/ preceded by /\textipa{aU}/ is not present in the Singapore corpus)}
    \label{fig:darkL_preceding_token_f1transition_comparison}
\end{figure}

In Figure \ref{fig:darkL_preceding_token_f1transition_comparison}, we perform a more detailed analysis by breaking down the syllable-final /\textipa{l}/ tokens according to their different preceding vowels and observe that this trend for F1 transition is generally consistent across most preceding vowels.

\subsubsection{Falling F2} 
\begin{figure}[t]
	\centering
	\includegraphics[width=\linewidth]{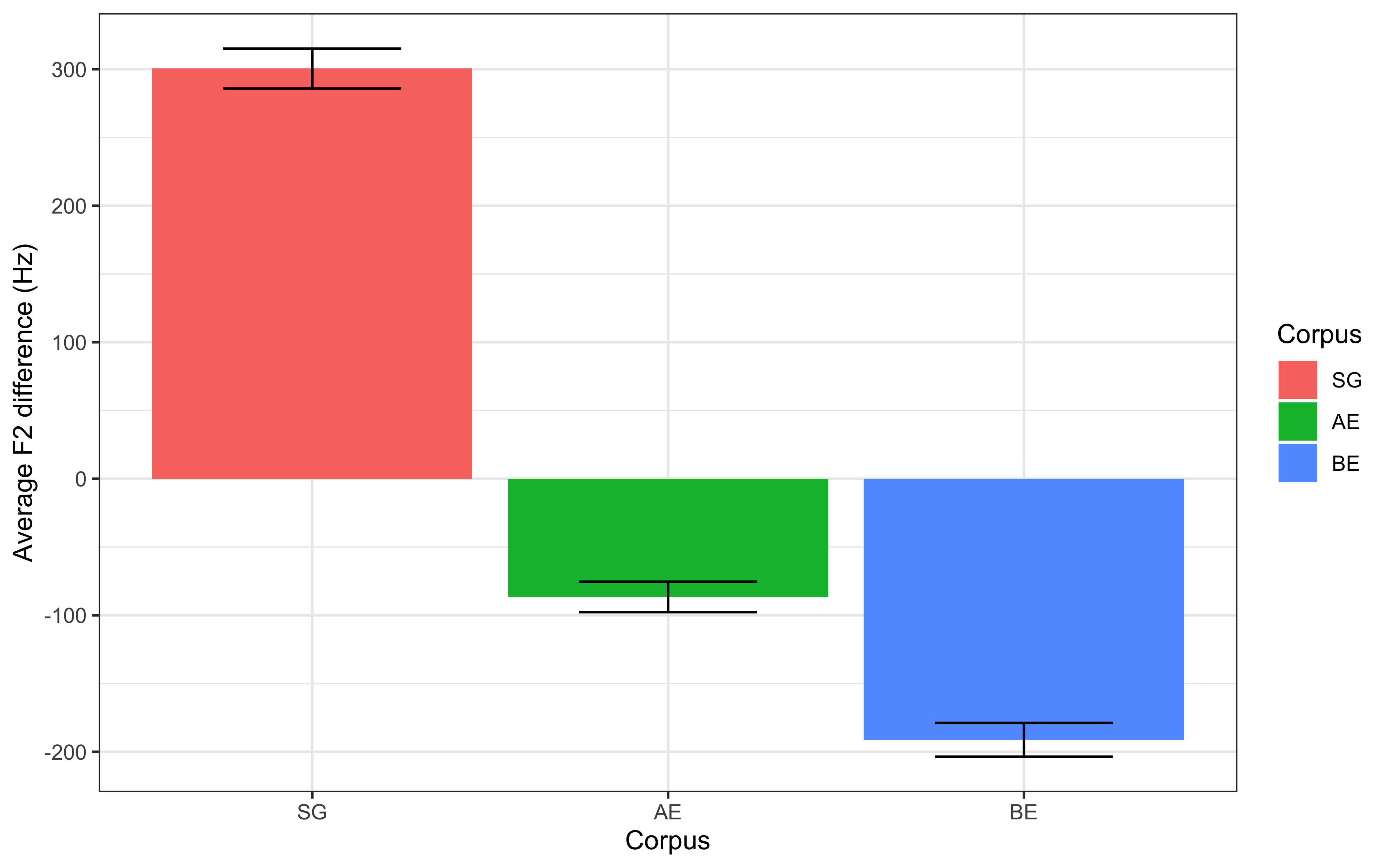}
    \caption{Comparison of F2 transition into syllable-final /\textipa{l}/. Singaporean children do not show conspicuous lowering of F2, compared to the other two speaker groups. The F2 difference plotted is calculated using the formula in Equation \ref{eq:F2Diff} on a per speaker basis. The error bars indicate standard error.}
    \label{fig:darkL_general_f2transition_comparison}
\end{figure}

In addition to falling F1, when transitioning into the syllable-final /\textipa{l}/ phoneme, we also anticipate a lowering of F2 as compared to that in the preceding vowel \cite{Johnson:11}. Similar to our analysis for falling F1, we plot the corresponding differences for F1 as in Figure \ref{fig:darkL_general_f2transition_comparison}, where the differences are calculated using Equation \ref{eq:F2Diff}.

\begin{equation}
    \begin{split} \label{eq:F2Diff}
        \text{F2 difference} = \text{Average F2 in syllable-final /\textipa{l}/} - \\ \text{Average F2 in preceding vowel}
    \end{split}
\end{equation}

Similar to the trend observed for F1 transition, syllable-final /\textipa{l}/ tokens produced by American (\textit{M} = -86.5) and British children (\textit{M} = -191) show a decrease in F1 estimates compared to its preceding token whereas Singaporean children demonstrated an increase (\textit{M} = 301). One-way ANOVA tests demonstrated that these differences are statistically significant across speaker groups, \textit{F}(2, 825) = 350.9, \textit{p} $<$ 0.001.  We then followed up our ANOVA test with a post hoc Tukey's HSD Test which showed that, the differences between Singaporean and American children speakers, as well as Singaporean and British children speakers are significant at \textit{p} $<$ 0.001. This analysis on the F2 values further affirmed our conclusion based on F1 values, showing that Singaporean children maybe producing something that is acoustically much less characteristic of a typical syllable-final /\textipa{l}/ compared to other speakers in terms of F2 characteristics as well. 

\begin{figure}[t]
	\centering
	\includegraphics[width=\linewidth]{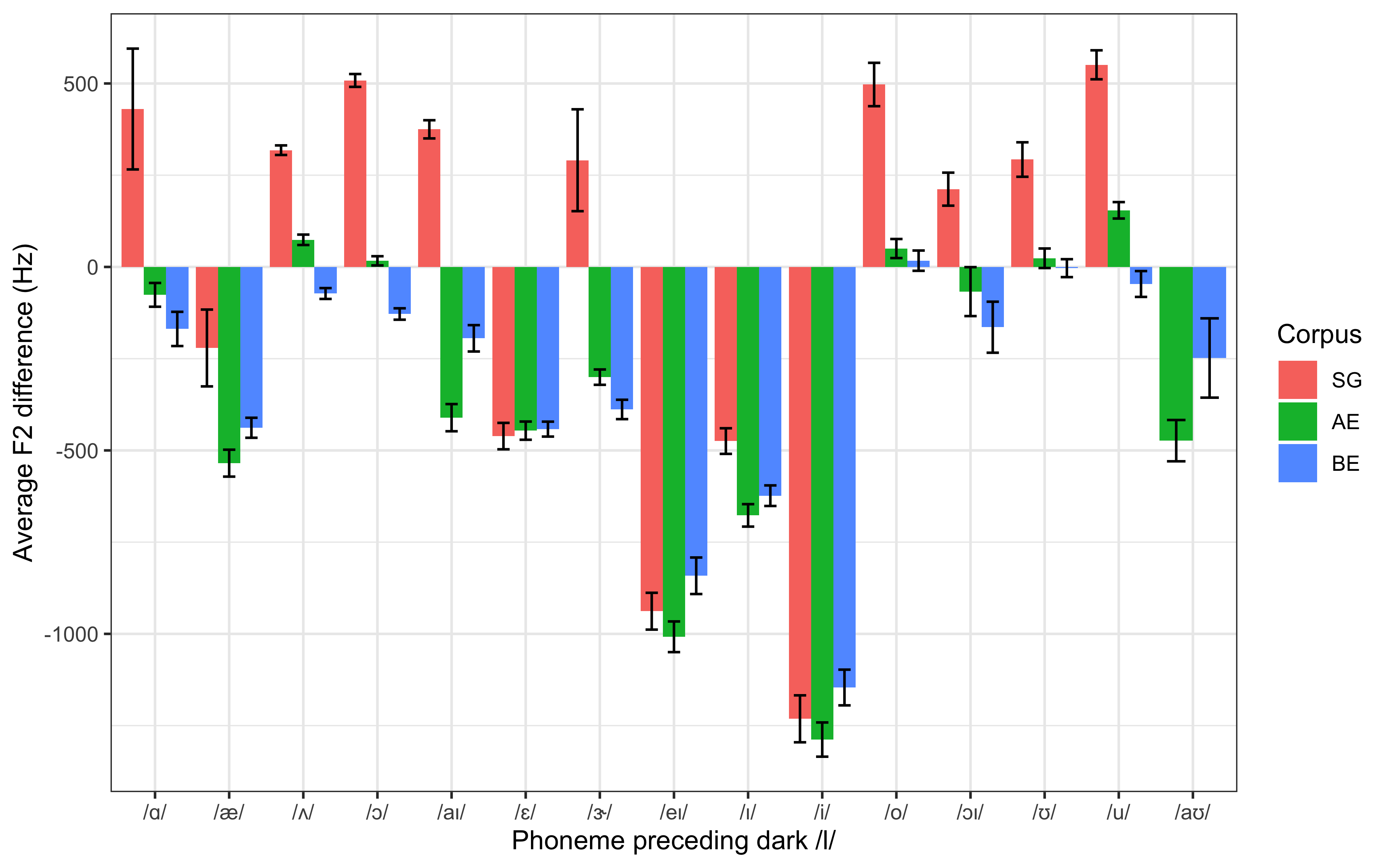}
    \caption{Syllable final /\textipa{l}/ F2 transition comparison by preceding phoneme across speaker groups. Compared to the other two speaker groups, Singaporean children generally do not demonstrate lowering of F2 as clearly in this transition. The error bars indicate standard error.(Syllable final /\textipa{l}/ preceded by /\textipa{aU}/ is not present in the Singapore corpus)}
    \label{fig:darkL_preceding_token_f2transition_comparison}
\end{figure}

We perform a more detailed analysis by breaking down the syllable-final /\textipa{l}/ tokens according their different preceding vowels in Figure \ref{fig:darkL_preceding_token_f2transition_comparison} and show that this trend for F2 transition (where Singaporean children do not exhibit falling F2 as clearly) is consistent across most preceding vowels.

\subsubsection{Gap between F2 and F3} 
\begin{figure}[t]
	\centering
	\includegraphics[width=\linewidth]{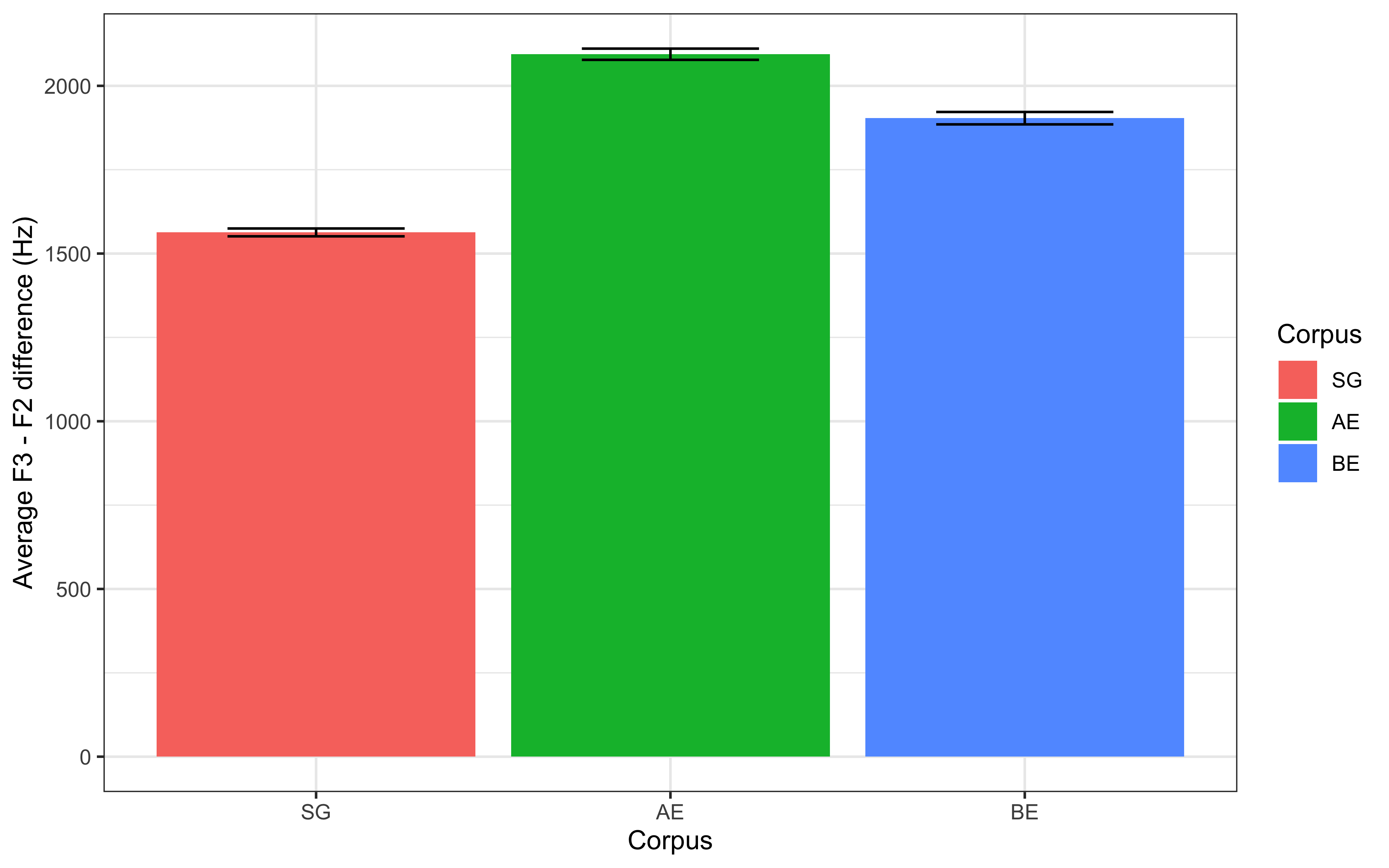}
    \caption{Comparison of F3 - F2 gap in syllable-final /\textipa{l}/. Singaporean children show a less conspicuous gap between F2 and F3, compared to the other two speaker groups. The F3 - F2 difference plotted is calculated using the formula in Equation \ref{eq:F3F2Diff} per token. The error bars indicate standard error.}
    \label{fig:darkL_general_f3-f2damping_comparison}
\end{figure}

In addition, if a phoneme is indeed articulated as a characteristic syllable-final /\textipa{l}/, we expect a relatively wide gap between F2 and F3 \cite{Liberman:88, Stevens:98}. We compare the difference between F2 and F3 values across different speaker groups in Figure \ref{fig:darkL_general_f3-f2damping_comparison}, which shows that this difference in formant values is the least in Singaporean children (\textit{M} = 1563) compared to American (\textit{M} = 2094) and British children (\textit{M} = 1904) who demonstrated more conspicuous gaps as should be prominent in syllable-final /\textipa{l}/. One-way ANOVA tests demonstrated that these differences are statistically significant across speaker groups, \textit{F}(2, 825) = 383, \textit{p} $<$ 0.001.  We then followed up our ANOVA test with a post hoc Tukey's HSD Test which showed that the differences in F3 - F2 gap is significant across all pairwise comparisons (\textit{p} $<$ 0.001). Therefore, although should be a conspicuous gap between F2 and F3 if a characteristic syllable-final /\textipa{l}/ was articulated, but this characteristic is significantly less clearly demonstrated by Singaporean children compared to the other two speaker groups.

\begin{figure}[t]
	\centering
	\includegraphics[width=\linewidth]{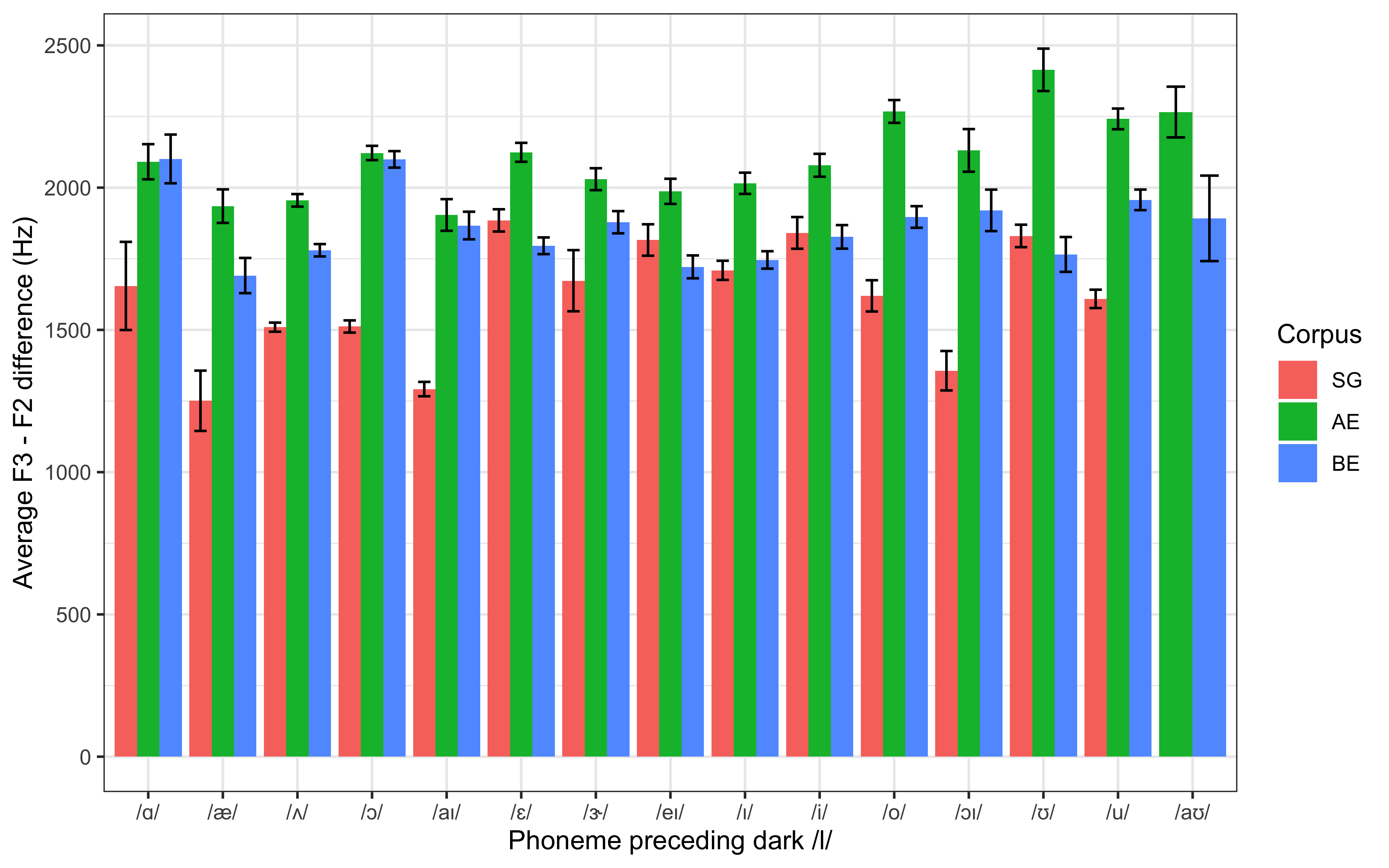}
    \caption{Syllable final /\textipa{l}/ F3 - F2 gap comparison by preceding phoneme across speaker groups. Compared to the other two speaker groups, Singaporean children generally show a less conspicuous gap between F2 and F3. The error bars indicate standard error. (Syllable final /\textipa{l}/ preceded by /\textipa{aU}/ is not present in the Singapore corpus)}
    \label{fig:darkL_preceding_token_f3-f2damping_comparison}
\end{figure}

In Figure \ref{fig:darkL_preceding_token_f3-f2damping_comparison}, we show that across the different preceding vowel conditions, this trend of Singaporean children showing a less conspicuous gap between F2 and F3 for syllable-final /\textipa{l}/ is consistent. 

This analysis on differences between F2 and F3 formant values further support what we have concluded using F1, F2 formants alone, that Singaporean children do not produce a characteristic syllable-final /\textipa{l}/ at where it occurs.

\begin{equation}
    \begin{split} \label{eq:F3F2Diff}
        \text{F3 - F2 difference} = \text{Average F3 in syllable-final /\textipa{l}/} - \\ \text{Average F2 in syllable-final /\textipa{l}/ }
    \end{split}
\end{equation}

\subsubsection{Difference between F1 and F2}
\begin{figure}[t]
	\centering
	\includegraphics[width=\linewidth]{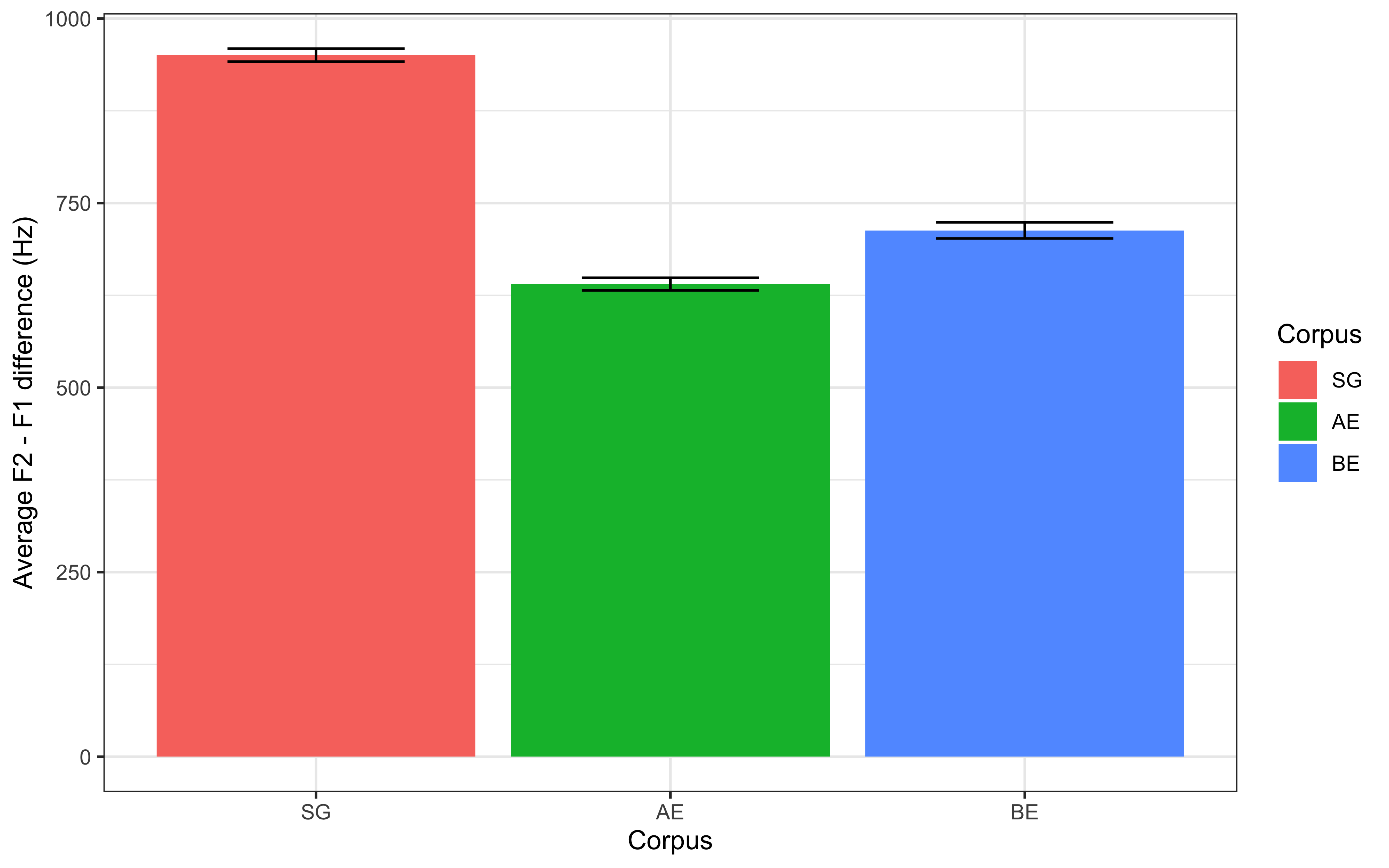}
    \caption{Comparison of F2 - F1 gap in syllable-final /\textipa{l}/. The F2 - F1 difference is the greatest in Singaporean children, where F1 and F2 should in fact be close together if the syllable-final /\textipa{l}/ phoneme was indeed present. The F2 - F1 difference plotted is calculated using the formula in Equation \ref{eq:F2F1Diff} per token. The error bars indicate standard error.}
    \label{fig:darkL_general_f2-f1damping_comparison}
\end{figure}

In syllable-final /\textipa{l}/, we also expect the difference between F1 and F2 values to be small \cite{Bayley:15}. The differences between F1 and F2 values across the speaker groups are visualized in Figure \ref{fig:darkL_general_f2-f1damping_comparison}, showing that such difference is the greatest in Singaporean children (\textit{M} = 950) where F1 and F2 should be close together if the syllable-final /\textipa{l}/ phoneme was indeed present, compared to American (\textit{M} = 640) and British children (\textit{M} = 713) who demonstrated smaller differences between F1 and F2 values. One-way ANOVA tests demonstrated that these differences are statistically significant across speaker groups, \textit{F}(2, 825) = 348.1, \textit{p} $<$ 0.001.  A post hoc Tukey's HSD Test showed that across all pairwise comparisons, F2 - F1 differences are significantly different across the speaker groups (\textit{p} $<$ 0.001). Therefore, although the difference between F1 and F2 should be small if a characteristic syllable-final /\textipa{l}/ was articulated, Singaporean children instead demonstrated significantly greater difference between the two formants compared to the other two speaker groups.

\begin{figure}[t]
	\centering
	\includegraphics[width=\linewidth]{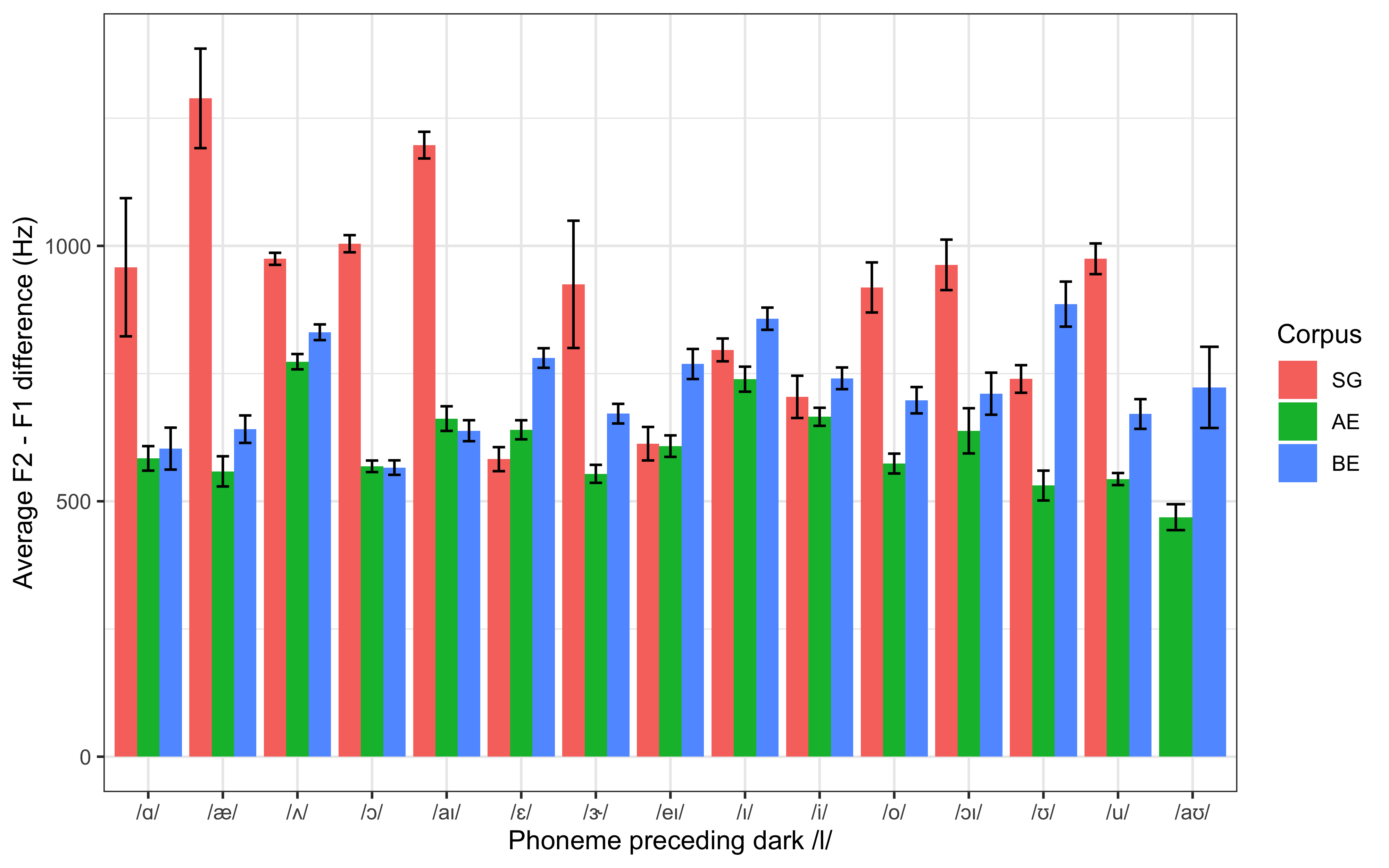}
    \caption{Syllable final  /\textipa{l}/ F2 - F1 gap comparison by preceding phoneme across speaker groups. F1 and F2 should be close together for the syllable-final /\textipa{l}/ phoneme, but Singaaporean children generally show the greatest F2 - F1 gap. The error bars indicate standard error. (Syllable final /\textipa{l}/ preceded by /\textipa{aU}/ is not present in the Singapore corpus)}
    \label{fig:darkL_preceding_token_f2-f1damping_comparison}
\end{figure}

In Figure \ref{fig:darkL_preceding_token_f2-f1damping_comparison}, we compare differences between F1 and F2 values for the three speaker groups across the various preceding vowel conditions. We observe that this trend of Singaporean children showing a greater difference between F1 and F2 for syllable-final /\textipa{l}/ is consistent across almost all preceding vowel conditions, when the difference is instead expected to be small for a characteristic syllable-final /\textipa{l}/. 

This analysis on differences between F1 and F2 formant values yields results that are consistent with what we have found using F1, F2 formants as well as analyzing the F2 and F3 gap. All four analyses agree and together, provide evidence for the observation that Singaporean children are not producing a characteristic syllable-final /\textipa{l}/ at where it occurs.

\begin{equation}
    \begin{split} \label{eq:F2F1Diff}
        \text{F2 - F1 difference} = \text{Average F2 in syllable-final /\textipa{l}/} - \\ \text{Average F1 in syllable-final /\textipa{l}/ }
    \end{split}
\end{equation}

\subsection{Summary: Singaporean children's  acoustic realization of syllable-final /\textipa{l}/ differs from British and American children, and from the canonical syllable-final /\textipa{l}/.}

From our acoustic analysis, when the phoneme /\textipa{l}/ is in the syllable-final position, Singaporean children tend to pronounce it differently compared to the other two speaker groups, giving it formant features differing from other speakers and from the canonical syllable-final /\textipa{l}/. According to the four acoustic characteristics that we have examined, the phonological interpretation would be that Singaporean children speakers' production of syllable-final /\textipa{l}/ tokens undergo extreme articulatory change differing from an characteristic syllable-final /\textipa{l}/ as reported in literature. That is, Singaporean children's syllable-final /\textipa{l}/ tokens, unlike the characteristic syllable-final /\textipa{l}/, are less prominent in showing characteristics of falling F1 and F2, conspicuous gap between F2 and F3, as well as small F2 - F1 difference.  From these observations, we conclude that Singaporean children's syllable-final /\textipa{l}/s could well be altered into a very different sound or considered to be deleted. This potentially explains the results from our unsupervised clustering experiments which show that syllable-final /\textipa{l}/ tokens produced by Singaporean children are mostly grouped into a different group than the majority group for American and British children. 

Given that studies like \citet{Hansen2001LinguisticCO} and \citet{He2014ProductionOE} have shown that speakers of Mandarin experienced difficulties in producing syllable-final /\textipa{l}/, the observation that Singaporean children's syllable-final /\textipa{l}/s are altered into a very different sound in terms of formant features could possibly be attributed to influence from the various Chinese languages spoken \cite{deterding2007ethnic} in Singapore.


\section{Analysis of rhotic approximant /\textipa{\*r}/}
The most characteristic acoustic feature of the rhotic approximant in English is that the third formant has very low frequency \cite{Stevens:98, Ladefoged:15}. Similar to our analysis on the lateral approximant /\textipa{l}/, we focus our analysis here on /\textipa{\*r}/ tokens that occur in syllable-final positions since Mandarin speakers tend to modify English syllable codas \citep{Hansen2001LinguisticCO}. For /\textipa{\*r}/, it is known that many Mandarin speakers drop the final /\textipa{\*r}/ in Mandarin \citep{Cheng1973} which could then potentially influence their pronunciation of the English syllable-final /\textipa{\*r}/. Given the various Chinese languages spoken in Singapore \citep{deterding2007ethnic}, we investigate if Singaporean children's pronunciation of the phoneme is similarly affected. For other two speaker groups, Standard BBC English is not rhotic whereas rhotic accents are the norm in most parts of North America \citep{Ladefoged:15}.

\subsection{Unsupervised clustering}
Archetypal analysis and Kmeans clustering using the characteristic F3 formant feature were carried out to explore potential differences across speaker groups.\footnote{We also tried the unsupervised clustering experiments using \textit{F1}(Hz), \textit{F2}(Hz), and \textit{F3}(Hz) estimates together. Those experiments gave results very similar to that of using just \textit{F3}(Hz) alone, therefore we only present the results from using just \textit{F3}(Hz) since that is the most characteristic formant feature for our analysis in this section.} Clustering results using the two unsupervised clustering methods are summarized in Tables \ref{tab:r_archetypal_clustering_F3} and \ref{tab:r_kmeans_clustering_F3} respectively. The two numbers in each row add up to 1.0 (stands for 100\%), illustrating the proportion of each speaker group's tokens that gets grouped into Group 1 and Group 2 respectively. Both sets of results suggest that when a syllable-final /\textipa{\*r}/ is present, Singaporean and British children are producing tokens with F3 characteristics differing from that of American children. We follow up on this result with a detailed acoustics analysis using the F3 formant estimates of syllable-final /\textipa{\*r}/ tokens.

\begin{table}[t]
  \begin{center}
    \small\addtolength{\tabcolsep}{-1pt}
	\begin{tabular}{|c|c|c|}
    \hline
   	\multirow{2}{*}{Corpus} & Group1 & Group2 \\
    &  3627 & 2021  \\
   	\hline
   	SG & \textbf{0.734} & 0.266 \\
   	AE & 0.014 & \textbf{0.986} \\
   	BE  & \textbf{0.793} & 0.207 \\
   	\hline
    \end{tabular}
    \caption{Archetypal analysis using \textit{F3}(Hz) estimates of syllable-final  /\textipa{\*r}/ from Singaporean, American and British children. Archetypal extreme points for each cluster are in terms of F3.}
    \label{tab:r_archetypal_clustering_F3}
  \end{center}
    
\end{table}

\begin{table}[t]
  \begin{center}
    \small\addtolength{\tabcolsep}{-1pt}
	\begin{tabular}{|c|c|c|}
     \hline
   	\multirow{2}{*}{Corpus} & Group1 & Group2 \\
    &  3000 & 2380  \\
   	\hline
   	SG & \textbf{0.880} & 0.120 \\
   	AE & 0.029 & \textbf{0.971} \\
   	BE  & \textbf{0.951} &  0.049 \\
   	\hline
    \end{tabular}
    \caption{Kmeans clustering using \textit{F3}(Hz) estimates of syllable-final  /\textipa{\*r}/ from Singaporean, American and British children. Centroids for each cluster are in terms of F3.}
    \label{tab:r_kmeans_clustering_F3}
  \end{center}
    
\end{table}

\subsection{Acoustic Analysis}

\begin{figure}[t]
	\centering
	\includegraphics[width=\linewidth]{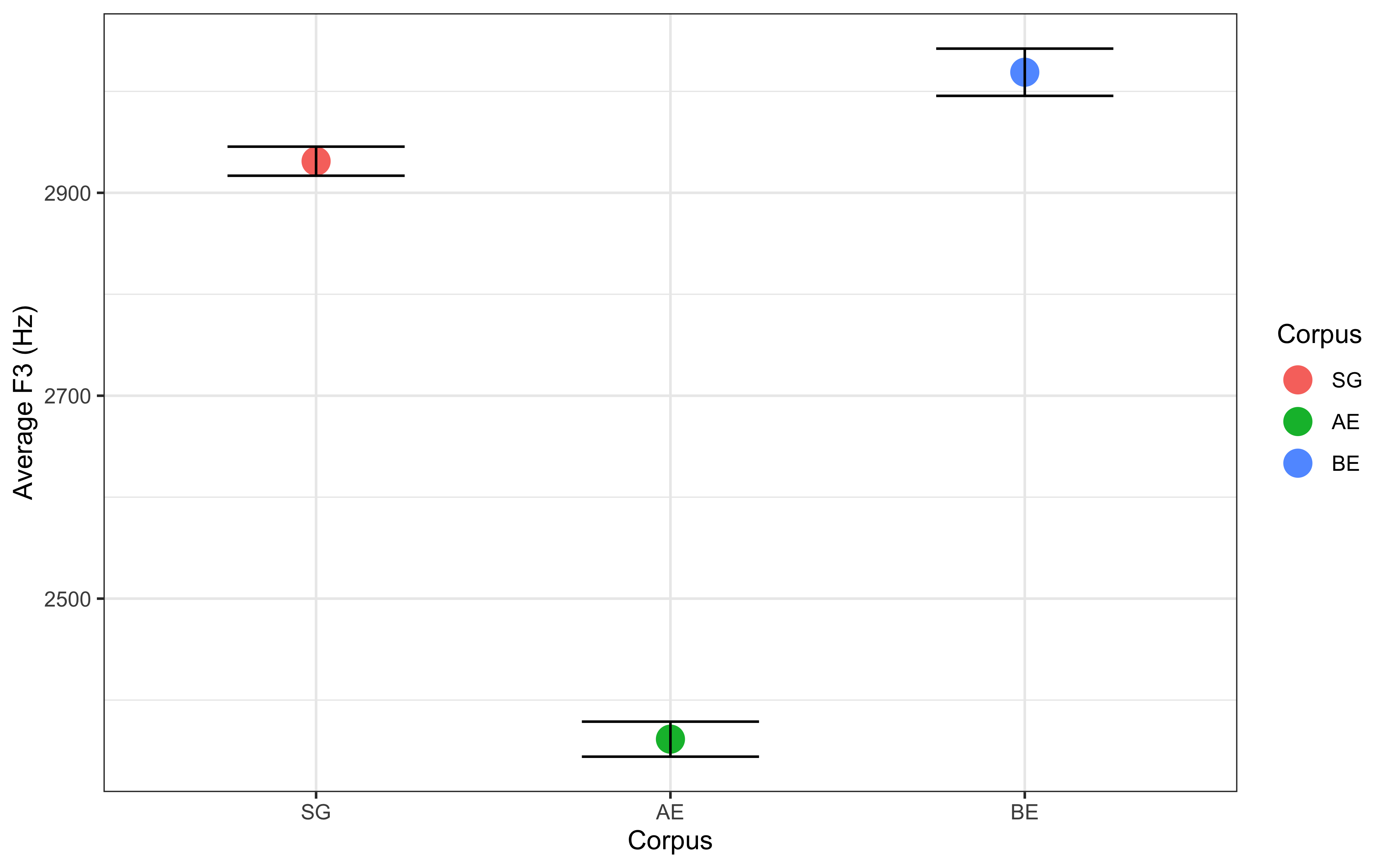}
    \caption{Syllable final /\textipa{\*r}/ F3 comparison across speaker groups. Singaporean and British children exhibit non-rhotic characteristics at syllable-final /\textipa{\*r}/ in contrast to American children. Colored points mark the mean F3 for each speaker group, whereas the error bars indicate standard error.}
    \label{fig:R_final_f3_comparison}
\end{figure}

We compared the mean F3 formant values across the three speaker groups where a syllable-final /\textipa{\*r}/ was supposed to be present. From the visualization of F3 formant values across speakers in Figure \ref{fig:R_final_f3_comparison}, we observe that syllable-final /\textipa{\*r}/ tokens produced by Singaporean (\textit{M} = 2931) and British children (\textit{M} = 3019) have much higher F3 formant values compared to that of American children (\textit{M} = 2361). An one-way ANOVA test demonstrated that these differences are statistically significant across speaker groups, \textit{F}(2, 411) = 404.9, \textit{p} $<$ 0.001. A post hoc Tukey's HSD Test further showed that the F3 estimates for American children are significantly lower than that of both Singaporean and British children (\textit{p} $<$ 0.001), but the difference between the latter two groups is not as significant (\textit{p} $=$ 0.00328). This aligns with our unsupervised clustering results, which also show that Singaporean and British children's syllable-final /\textipa{\*r}/ have F3 characteristics differing from that of American children.

\subsection{Summary: Singaporean and British children exhibit non-rhotic characteristics at syllable-final /\textipa{\*r}/ in contrast to American children}
Our unsupervised clustering results show that for syllable-final /\textipa{\*r}/, Singaporean and British children have F3 characteristics differing from that of American children. The follow up acoustic analysis further elucidates this difference, showing that Singaporean and British children speakers do not lower their F3 formant values nearly as much as American children speakers do for syllable-final /\textipa{\*r}/. In terms of articulatory implications, this suggests that Singaporean and British children speakers lack rhoticity in their production of syllable-final /\textipa{\*r}/ tokens, in contrast to American children.

\section{Discussion}
The vowel height and frontness characteristics that different speaker groups exhibit are summarized in Tables \ref{tab:tongue_height_overall_table} and \ref{tab:tongue_frontness_overall_table}. We observe that the height and frontness of vowels produced by Singaporean children often exhibit characteristics that are opposite from that of British children. For instance, Singaporean children produce TRAP$-$BATH split vowels, /\textipa{\ae}/ and /\textipa{E}/ all with a higher tongue position, while British children demonstrate lower tongue height for these vowels and higher tongue height for the rest of the vowels examined. British children also demonstrate fronting of vowels for the only vowels that Singaporean children do not show fronting for, namely, /\textipa{u}/ and /\textipa{U}/. In terms of fronting of the vowels, Singaporean children's behavior appears more similar compared to American children. However, vowel height characteristics demonstrated by Singaporean children does not resemble any of the other two populations. 

\begin{table}[t]
  \begin{center}
  \small\addtolength{\tabcolsep}{-2pt}
  \begin{tabular}[\linewidth]{|c|p{1.15cm}|p{1.15cm}|c|c|c|c|c|c|c|c|}
   \hline
    \multirow{3}{*}{Corpus} & \multicolumn{10}{c|}{Vowels and tongue height}\\
	\hhline{~----------}
	   & \multicolumn{2}{c|}{TRAP$-$BATH} &  \multirow{2}{*}{/\textipa{\ae}/} &  \multirow{2}{*}{/\textipa{E}/} & \multirow{2}{*}{/\textipa{A}/} & \multirow{2}{*}{/\textipa{O}/} & \multirow{2}{*}{/\textipa{u}/} & \multirow{2}{*}{/\textipa{U}/} & \multirow{2}{*}{/\textipa{i}/} & \multirow{2}{*}{/\textipa{I}/} \\
	   	 \hhline{~--~~~~~~~~}
	   	&  [\textipa{\ae}] & [\textipa{A}]  & & & & & & & &  \\

    \hline
	SG & \cellcolor{black} H & \cellcolor{black} & \cellcolor{black} &  \cellcolor{black}  &  & &  & \cellcolor{gray} &  &  \cellcolor{black}\\
      \hline
      AE &  & \cellcolor{black} & &  &  & & \cellcolor{gray} & & \cellcolor{black} & \\
      \hline
      BE &  &  & & & \cellcolor{black} & \cellcolor{black} & \cellcolor{black} & \cellcolor{black} & \cellcolor{black} &  \cellcolor{gray} \\
	\hline
  \end{tabular}
  \caption{Summary of tongue height characteristics exhibited by speaker groups for different vowels. Darker boxes indicate that the speaker group exhibits higher tongue positions for the particular vowel compared to other speaker groups.}
  \label{tab:tongue_height_overall_table}
  \end{center}
 \end{table}

\begin{table}[t]
  \begin{center}
  \small\addtolength{\tabcolsep}{-2pt}
  \begin{tabular}[\linewidth]{|c|p{1.15cm}|p{1.15cm}|c|c|c|c|c|c|c|c|}
   \hline
    \multirow{3}{*}{Corpus} & \multicolumn{10}{c|}{Vowels and tongue frontness}\\
	\hhline{~----------}
	   & \multicolumn{2}{c|}{TRAP$-$BATH} &  \multirow{2}{*}{/\textipa{\ae}/} &  \multirow{2}{*}{/\textipa{E}/} & \multirow{2}{*}{/\textipa{A}/} & \multirow{2}{*}{/\textipa{O}/} & \multirow{2}{*}{/\textipa{u}/} & \multirow{2}{*}{/\textipa{U}/} & \multirow{2}{*}{/\textipa{i}/} & \multirow{2}{*}{/\textipa{I}/} \\
	   	 \hhline{~--~~~~~~~~}
	   	& [\textipa{\ae}]&[\textipa{A}]& & & & & & & &  \\

    \hline
	SG & \cellcolor{black}   & \cellcolor{gray}  &  \cellcolor{black}  &  \cellcolor{black}  & \cellcolor{black}  & \cellcolor{gray} &  &  &  \cellcolor{black} &  \cellcolor{black}\\
      \hline
      AE & \cellcolor{gray}  & \cellcolor{black}  &  \cellcolor{black}  &  \cellcolor{gray}  & \cellcolor{gray}  & \cellcolor{black} & \cellcolor{gray} & \cellcolor{gray} & \cellcolor{black} & \cellcolor{gray} \\
      \hline
      BE &  &   & & &  &  & \cellcolor{black} & \cellcolor{black} &  &  \\
	\hline
  \end{tabular}
  \caption{Summary of tongue frontness characteristics exhibited by speaker groups for different vowels. Darker boxes indicate that the speaker group exhibits more fronted tongue positions for the particular vowel compared to other speaker groups. }
  \label{tab:tongue_frontness_overall_table}
  \end{center}
 \end{table}
 
 
In tense and lax vowel pairs, /\textipa{u}/ and /\textipa{U}/, as well as /\textipa{i}/ and /\textipa{I}/, although all speakers show significant duration difference between the tense and lax vowels, we also consistently observe that the difference is less conspicuous for Singaporean children compared to the other speaker groups. For rhotic approximant /\textipa{\*r}/, Singaporean children are similar to British children in their lack of rhoticity. In the case of lateral approximant /\textipa{l}/, Singaporean children's syllable-final /\textipa{l}/ exhibits characteristics unlike either American or British children, and differs from a canonical syllable-final /\textipa{l}/.
 
\begin{figure}
    \centering
    \begin{subfigure}{0.8\linewidth}
    \centering
    \includegraphics[width=0.85\linewidth]{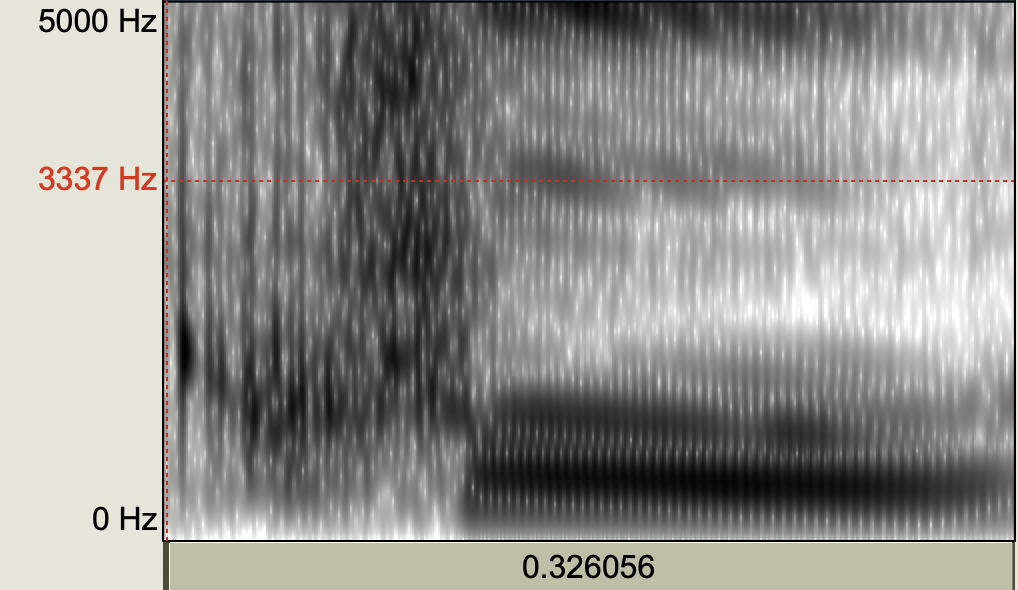}
    \vspace{-1.5mm}
    \caption{Spectrogram example for syllable-final /\textipa{l}/ token produced by a Singaporean child as in the word ``cool" (/\textipa{kul}/).}
    \vspace{2mm}
    \label{fig:SG_L_spectrogram}
    \end{subfigure}
    \begin{subfigure}{0.8\linewidth}
    \centering
    \includegraphics[width=0.85\linewidth]{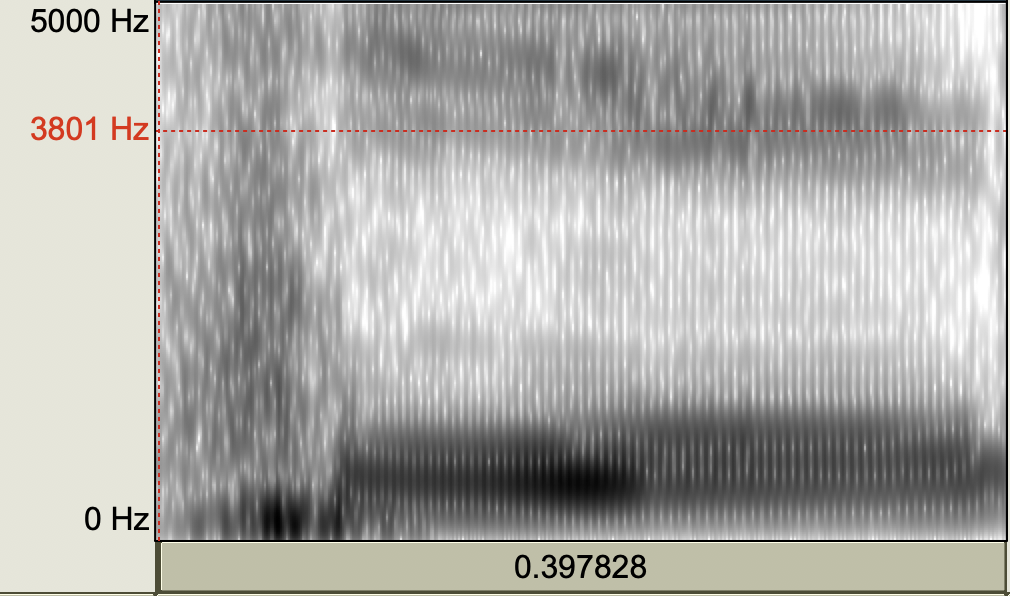}
    \vspace{-1.5mm}
    \caption{Spectrogram example for syllable-final /\textipa{l}/ token produced by an American child as in the word ``cool" (/\textipa{kul}/).}
    \vspace{2mm}
    \label{fig:AE_L_spectrogram}
    \end{subfigure}
    \begin{subfigure}{0.8\linewidth}
    \centering
    \includegraphics[width=0.85\linewidth]{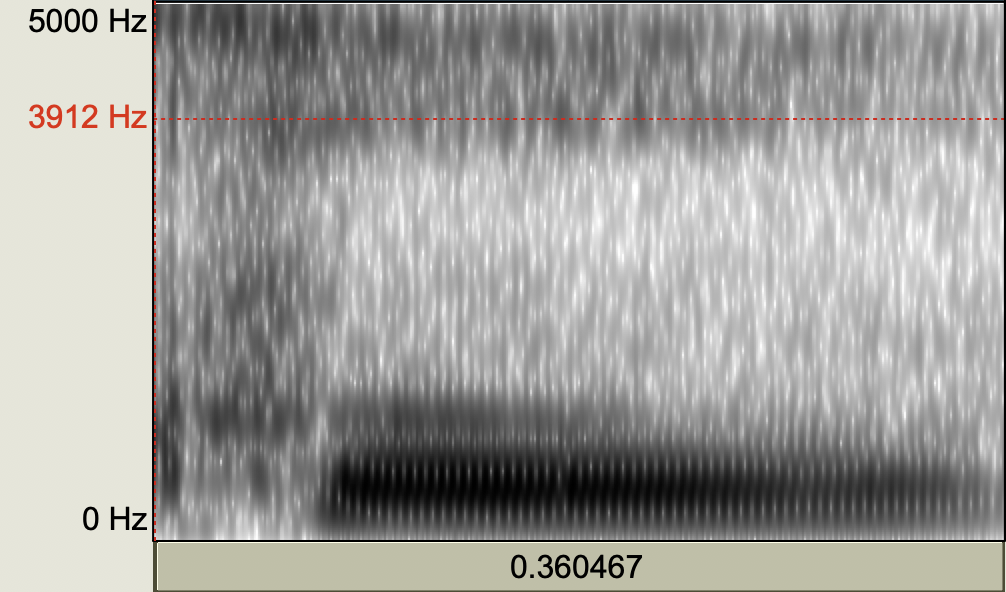}
    \vspace{-1.5mm}
    \caption{Spectrogram example for syllable-final /\textipa{l}/ token produced by a British child as in the word ``cool" (/\textipa{kul}/).}
    \label{fig:BE_L_spectrogram}
    \end{subfigure}
    \vspace{-3mm}
    \caption{Comparing spectrogram examples for syllable-final /\textipa{l}/ token produced by Singaporean, American and British children.}
\end{figure}

We show illustrative examples of syllable-final /\textipa{l}/ using instances of the word ``cool" (/\textipa{kul}/) pronounced by Singaporean, American and British children in Figures \ref{fig:SG_L_spectrogram}, \ref{fig:AE_L_spectrogram} and \ref{fig:BE_L_spectrogram} respectively. Comparing the formants in these spectrograms, American and British children's F1 and F2 for syllable-final /\textipa{l}/ are low and close together, but this is less so for Singaporean children whose /\textipa{l}/ tokens have F1 and F2 values being further apart. Further, one characteristic feature of syllable-final /\textipa{l}/ is that there is a huge gap between F2 and F3, but for Singaporean children F3 is closer towards being in the middle of F2 and F4, making the F2 and F3 gap less conspicuous. Specifically, in the spectrogram examples, F3 values for American and British children are around 3800Hz and 3900Hz whereas the F3 for the Singaporean speaker example is much lower (around 3300Hz) and therefore closer to the F2 formants.

We also performed preliminary analysis looking into other aspects such as prosody, as well as analyzing differences between spontaneous versus elicit speech across speakers. Such analysis revealed some interesting trends such as Singaporean children showing greater variation in their /\textipa{\ae}/ vs. /\textipa{E}/ vowels produced in spontaneous speech when compared to elicit speech. However, with these analysis, no consistent trends were observed across speaker groups nor for each vowel or approximant type. Therefore, in this paper, we focus on presenting our findings on formant and duration characteristics which revealed more consistent trends. Nonetheless, looking further into other aspects to analyze Singaporean children's pronunciation patterns would definitely be an interesting direction for future work. 

From our analysis in this work, we also found several pronunciation characteristics exhibited by Singaporean children that are not present for either American or British children. This work alludes to sociolinguistic perspectives of how Singapore English could have evolved beyond the British influence during historical colonization \cite{lim2015contact}, moving towards also embodying American pronunciation characteristics, and even traits beyond American and British pronunciation patterns. It would be interesting to compare Singaporean children's English to speech from other speaker groups in future work to study the different sources of influence. Investigating how Singapore English is also influenced by Malay \cite{zuraidah1997MalayEnglish} and the range of Chinese languages spoken \cite{deterding2007ethnic} in Singapore can paint a more comprehensive picture of the complexities of Singapore English; this is a line of on-going research endeavors.

\section{Conclusion}

We presented a large-scale study to characterize Singaporean children's English pronunciation patterns. We looked into different vowels (TRAP$-$BATH split vowels, /\textipa{\ae}/ vs. /\textipa{E}/, /\textipa{A}/ vs. /\textipa{O}/, /\textipa{u}/ vs. /\textipa{U}/, and /\textipa{i}/ vs. /\textipa{I}/) and approximants (/\textipa{l}/ and /\textipa{\*r}/) by first using unsupervised clustering to explore potential trends across the different speaker groups, followed by detailed acoustic analysis with linguistics insights. Our analysis showed that Singaporean children speakers are generally more similar to American children speakers in their pronunciation patterns of the vowels compared to British speakers, in exhibiting fronting. Whereas, when it comes to the pronunciation of syllable-final /\textipa{\*r}/, Singaporean children speakers, like British children speakers, demonstrated a lack of rhoticity. It is also interesting that Singaporean children speakers appeared to have a different pronunciation for the approximant /\textipa{l}/ distinct from the other speaker groups.

Our work is the first of the its kind to characterize Singapore English pronunciation patterns by quantifying pronunciation differences across different English speaker groups on a large-scale basis, and to present a comparative study from the perspective of children speech.
Given the multicultural and multilingual environment in Singapore, Singapore English could potentially have been shaped by British influence during historical colonization, American influence through media sources, as well as bear other characteristics, for instance, originating from different languages spoken in Singapore such as Malay and Chinese. Future work analyzing Singaporean children's pronunciation patterns from other aspects and investigation into effects of the various influences will help to paint a more comprehensive picture of the complexities of Singapore English.






\clearpage
\bibliographystyle{elsarticle-num-names}
\bibliography{journal}







\end{document}